%% file: main_plus_sup.tex
\documentclass[10pt,twocolumn,letterpaper]{article}

\usepackage{wacv}
\usepackage{times}
\usepackage{epsfig}
\usepackage{graphicx}
\usepackage{amsmath}
\usepackage{amssymb}
\usepackage{booktabs}
\usepackage[acronym,nomain]{glossaries}
\input{abbreviation}
\usepackage{multirow}
\usepackage{tabularx}
\usepackage{textcomp, gensymb}

\DeclareMathOperator*{\argmin}{arg\,min}
\usepackage{algorithmic}	
\usepackage{algorithm}
\usepackage{subfigure}
\usepackage{enumitem}
\usepackage{caption}
\usepackage[accsupp]{axessibility}

\input{configs/definations}
\def\wacvPaperID{252} 
\wacvalgorithmstrack   
\wacvfinalcopy 

\ifwacvfinal
\usepackage[breaklinks=true,bookmarks=false]{hyperref}
\else
\usepackage[pagebackref=true,breaklinks=true,colorlinks,bookmarks=false]{hyperref}
\fi

\def\toptitlebar{
  \hrule height4pt
  \vskip .25in
}

\def\bottomtitlebar{
  \vskip .25in
  \hrule height1pt
  \vskip .25in
}

\begin{document}

\title{\ourTitle}
\author{
Yifan Lu \\
ETH Zurich
\and
Gurkirt Singh \\
ETH Zurich
\and
Suman Saha \\
ETH Zurich
\and
Luc Van Gool \\
ETH Zurich, KU Leuven
}

\maketitle
\thispagestyle{empty}

\input{text/abstract}
\input{text/intro}
\input{text/contrib}
\input{text/related_work}
\input{text/method}
\input{text/experiments}

\input{text/conclusion}

\onecolumn
\setlength{\parindent}{0em}

\hsize\textwidth\linewidth
\hsize\toptitlebar 
{
  \centering
  {
    \Large
    \bfseries 
    Exploiting Instance-based Mixed Sampling via \\
    Auxiliary Source Domain Supervision for Domain-adaptive Action Detection\\
    \emph{Supplementary Materials}
    \par
  }
}
\bottomtitlebar 

\input{text/sup-mat/dataset_creation}


\par\vfill\par
\clearpage

{\small
\bibliographystyle{ieee_fullname}
\bibliography{references}
}

\end{document}

%% file: abbreviation.tex
\newacronym{daaim}{DA-AIM}{domain-adaptive action instance mixing}
\newacronym{gan}{GAN}{generative adversarial nets}
\newacronym{cnn}{CNNs}{convolutional neural networks}
\newacronym{uda}{UDA}{unsupervised domain adaptation}
\newacronym{da}{DA}{domain adaptation}
\newacronym{dacs}{DACS}{Domain Adaptation via Cross-domain Mixed Sampling}
\newacronym{grl}{GRL}{gradient reversal layer}
\newacronym{roi}{RoI}{Region-of-Interest}
\newacronym{ccw}{ccw}{counter-clockwise}
\newacronym{mlp}{MLP}{multilayer perceptrons}
\newacronym{sgd}{SGD}{Stochastic Gradient Descent}
\newacronym{map}{mAP}{Mean Average Precision}
\newacronym{rnn}{RNNs}{Recurrent Neural Networks}
\newacronym{i3d}{I3D}{inflated 3D CNN}
\newacronym{rcnn}{R-CNN}{Region Convolution Neural Network}
\newacronym{tcnn}{T-CNN}{Tube Convolutional Neural Network}
\newacronym{mmd}{MMD}{Maximum Mean Discrepancy}
\newacronym{mdd}{MDD}{Margin Disparity Discrepancy}
\newacronym{ssl}{SSL}{Semi-Supervised Learning}

%% file: configs/definations.tex
\newcommand{\qheading}[1]{\noindent\textbf{#1}}

\newcommand{\ourTitle}[0]{Exploiting Instance-based Mixed Sampling via \\ Auxiliary Source Domain Supervision for Domain-adaptive Action Detection}

\newcommand{\slowfast}[0]{\mbox{SlowFast}\xspace}
\newcommand{\pyslowfast}[0]{py\slowfast}

\newcommand{\avakin}[0]{\mbox{AVA-Kinetics}\xspace}
\newcommand{\avakins}[0]{\mbox{AVA-Kin}}

\newcommand{\sotalong}[0]{state-of-the-art\xspace}
\newcommand{\Sotalong}[0]{State-of-the-art\xspace}

\newcommand{\st}[0]{\mbox{spatiotemporal}\xspace}

\newcommand{\uda}[0]{\mbox{UDA}\xspace}
\newcommand{\udalong}[0]{\mbox{unsupervised domain adaptatio}\xspace}

\newcommand{\multisports}[0]{\mbox{MultiSports}\xspace}

\usepackage{pifont}
\newcommand{\cmark}{\color{green}\ding{51}}
\newcommand{\cmarkb}{\color{black}\ding{51}}
\newcommand{\xmark}{\color{red}\ding{55}}

\newcommand{\ihdone}[0]{\mbox{IhD-1}\xspace}
\newcommand{\ihdtwo}[0]{\mbox{IhD-2}\xspace}

%% file: text/abstract.tex
\begin{abstract}
We propose a novel domain adaptive action detection approach and a new adaptation protocol that leverages the recent advancements in image-level \gls{uda} techniques and handle vagaries of instance-level video data. Self-training combined with cross-domain mixed sampling has shown remarkable performance gain in semantic segmentation in UDA (unsupervised domain adaptation) context. Motivated by this fact, we propose an approach for human action detection in videos that transfers knowledge from the source domain (annotated dataset) to the target domain (unannotated dataset) using mixed sampling and pseudo-label-based self-training. The existing \gls{uda} techniques follow a ClassMix algorithm for semantic segmentation. However, simply adopting ClassMix for action detection does not work, mainly because these are two entirely different problems, i.e., pixel-label classification vs. instance-label detection. To tackle this, we propose a novel action instance mixed sampling technique that combines information across domains based on action instances instead of action classes. Moreover, we propose a new UDA training protocol that addresses the long-tail sample distribution and domain shift problem by using supervision from an auxiliary source domain (ASD). For the ASD, we propose a new action detection dataset with dense frame-level annotations. We name our proposed framework as \gls{daaim}. We demonstrate that \gls{daaim} consistently outperforms prior works on challenging domain adaptation benchmarks. The source code and datasets are available at \url{https://github.com/wwwfan628/DA-AIM}.
\end{abstract}

%% file: text/intro.tex
\begin{figure}[htbp!]
\begin{center}
   \includegraphics[width=0.8\columnwidth]{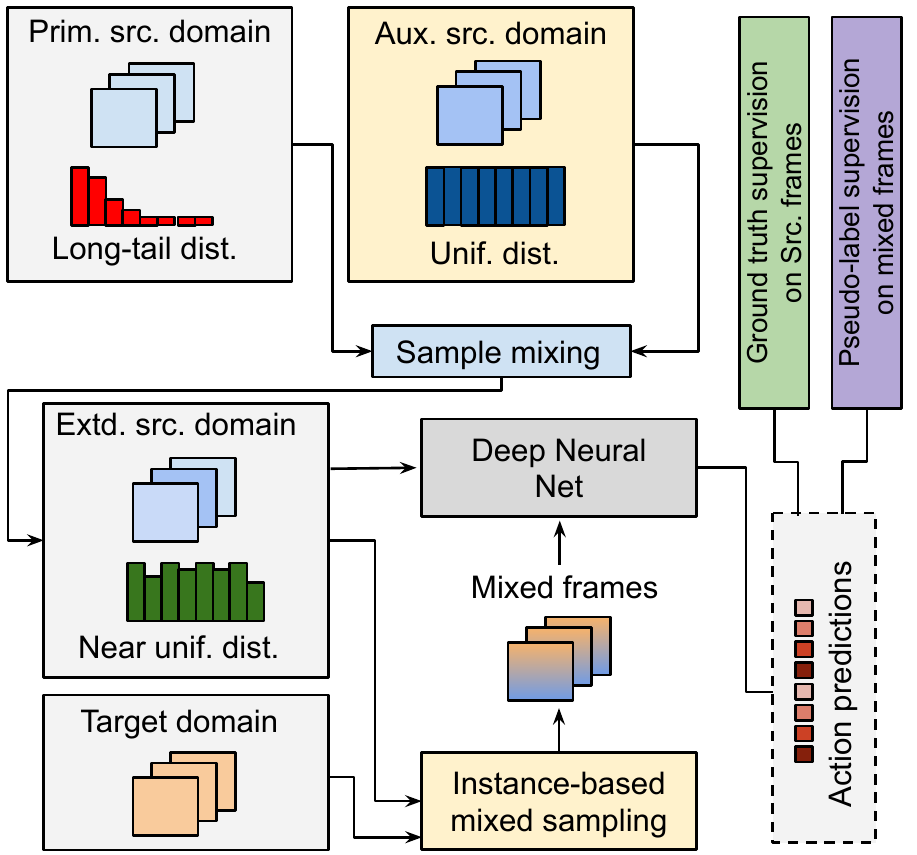}
\end{center}
\caption{\small{The above diagram illustrates the two main contributions of this work.
Firstly, We propose a novel instance-based cross-domain mixed sampling technique designed explicitly for video-based action detection.
Unlike the prior UDA method \cite{tranheden2021dacs}, which follows a class-based mixed sampling to generate augmented mixed images,
our mixed sampling algorithm randomly samples image patches based on the number of action instances present
in the source frames. The output is a set of mixed frames containing instances of the source and target domains.
Secondly, we propose to mix auxiliary source domain samples with the primary source domain to create a new extended source domain. 
This is done to address various problems such as long-tail distribution of the primary source domain, large variability in action instances across domains.
}}
\label{fig:introteaser}
\end{figure}

\section{Introduction} \label{sec:intro}
Over the past few years, we have witnessed tremendous progress in vision-based action detection\cite{kumar2022end,rana2021we,yang2019step,duarte2018videocapsulenet,li2018recurrent,ali2018real,singh2018tramnet,singh2017online,saha2017amtnet,behl2017incremental,hou2017tube,kalogeiton2017action,saha2016deep,weinzaepfel2015learning,tuber2022zhao,pan2021actor}. This success is largely attributed to the deep neural networks, which demonstrates superior performance in several computer vision tasks. However, these networks require expensive ground truth annotations to be trained appropriately under a supervised-learning setup. 
Particularly, for action detection, it is highly time-consuming and labor-intensive to generate such a large amount of annotated data~\cite{Li2020avakin,ava2018gu,li2021multisports,soomro2012ucf101}.
The main reason is that ground truth labels for both action categories and instances are required, i.e. all the action instances in a video frame need to be spatially localized using bounding boxes and these boxes are to be labeled with their respective action categories. As the video duration, the number of videos and action instances increase, the annotation cost rises rapidly, making the labeling process highly impractical and expensive.
One standard approach to circumvent this issue is to rely on unsupervised domain adaptation (UDA)~\cite{tranheden2021dacs,long2015learning,long2017deep,ganin2015unsupervised,ganin2016domain,li2016revisiting,li2018adaptive,saito2018maximum} in which knowledge transfer is performed by adapting the network trained on the source domain to the target domain.
The source domain refers to either synthetic data \cite{richter2016playing,ros2016synthia}
or publicly available real data \cite{ros2016synthia,Li2020avakin}
for which the ground truth annotations are available.
The target domain refers to real data for which ground truths are not accessible.

Prior works \cite{davar2011domain,zhu2013enhancing,liu2019deep,pan2020adversarial,song2020modality,
zhang2021progressive,yang2022interact,chen2020action,munro2020multi,choi2020shuffle,chen2019temporal}
mostly focus on domain-adaptive (DA) action recognition which is a simpler problem than DA action detection 
as the former requires only to solve the action classification without considering the much harder 
instance localization problem.
Agarwal \etal \cite{agarwal2020unsupervised} propose a DA action detection approach in which domain alignments of spatial and temporal features are performed using GRLs \cite{ganin2015unsupervised}.
They introduce two UDA benchmarks which are limited to only three/four sports actions.
Since there is no standard UDA benchmark available for action detection,
they rely on the sports-related action classes, which are common across different datasets (or domains).
Moreover, the datasets used in \cite{agarwal2020unsupervised} have low video resolution and are outdated.

In this work, we propose a generic UDA framework that is not limited to certain action categories and can be used for a larger set of action classes, \eg AVA~\cite{ava2018gu}.
First, we consider the train set from the \avakin~\cite{Li2020avakin} dataset as our primary source domain.
Since \avakin is a large-scale and diversified action detection dataset from YouTube videos,
using it as the source domain would allow the model to learn meaningful \st
representation and better adaption to the target domain.
However, it imposes two main challenges.
Firstly, \avakin has a long-tailed label distribution which biases the model towards 
certain action categories, resulting in a poor adaptation of under-represented classes.
Secondly, there is a large variability in actions (belonging to same action classes) across domains
due to factors like differences in capturing devices, backgrounds, 
temporal motion patterns, appearance. 
To tackle these problems, we propose to supervise the network 
using labeled training samples from an auxiliary source domain (ASD) (Fig. \ref{fig:introteaser}).
ASD alleviates the aforementioned problems by:
(a) injecting training samples of under-represented or missing classes into the source domain, and (b) recreating the action scenes to resemble the target domain scenes.
For ASD, we create a new action detection dataset with dense ground truth annotations.

We empirically found that the GRL-based approach (similar to \cite{agarwal2020unsupervised}) does not show any noticeable improvements in either of our UDA settings (\S \ref{sec:sota}). 
Recently, Tranheden \etal \cite{tranheden2021dacs} proposed a UDA method for semantic segmentation, which exhibits superior performance in semantic segmentation task.
Their method generates augmented training images following a cross-domain mixed sampling (CDMS) technique.
CDMS is suitable for pixel-level prediction (or segmentation) tasks. 
However, for instance-level (or bounding-box) prediction like action detection,
CDMS fails to generate meaningful training samples since these two are entirely different problems, i.e., pixel-label classification vs. instance-label detection.
To tackle this issue, we propose a novel action-instance-based mixed sampling technique that combines information across domains based on action instances present in the source domain.
For source-to-target knowledge transfer, we adapt the Mean Teacher based self-training \cite{tarvainen2017mean}. 
We name our proposed UDA framework as DA-AIM (\textbf{d}omain-\textbf{a}daptive \textbf{a}ction \textbf{i}nstance \textbf{m}ixing) (Fig. \ref{fig:introteaser}).
We are the first to propose a DA action detection framework based on cross-domain mixed sampling and self-training. We implement and compare with three state-of-the-art approaches and achieve best results on different \uda benchmarks.

\begin{figure*}[htbp!]
\begin{center}
\includegraphics[width=1.0\linewidth]{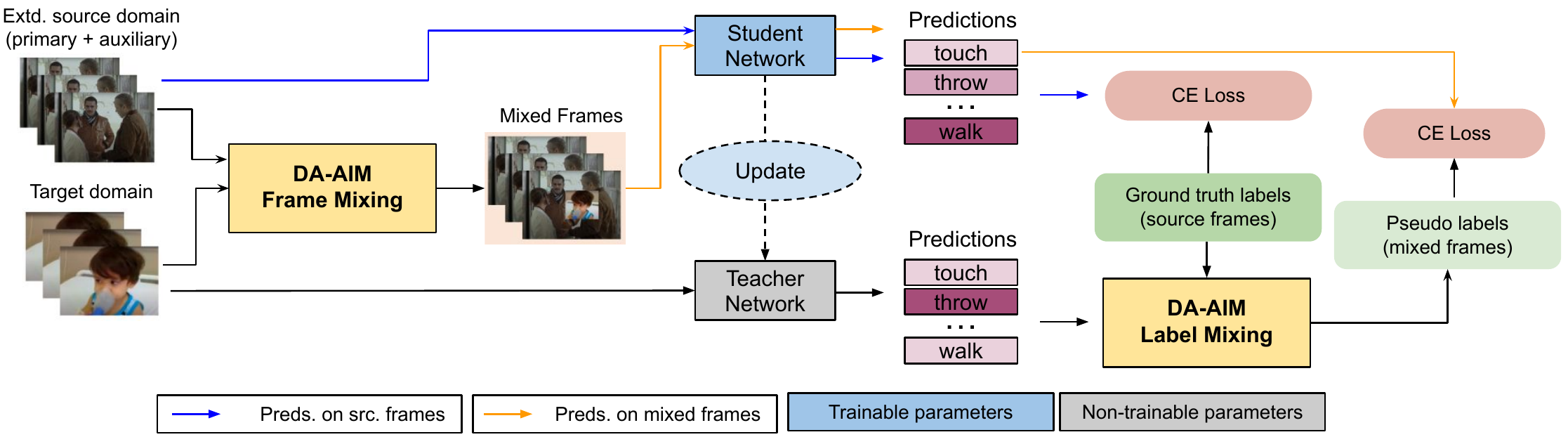}
\end{center}
   \caption{\small{Overview of the proposed DA-AIM framework.
      The basic building blocks of DA-AIM are (a) training sample mixing,
   (b) frame mixing, (c) label mixing, and (d) self-training.
   (a) We first generate an extended (extd.) source domain by mixing training examples of the primary and auxiliary source domains.
   (b) Next, the frame mixing module generates augmented video frames (or mixed frames) by mixing action instances of the source frame with the target frames. During mixing, spatial and temporal information are considered due to the inherent spatiotemporal nature of actions. The source and mixed frames are then fed to a deep neural network (called the student network). The student network is optimized with action classification losses. Ground truth labels are used to penalize wrong predictions on source frames, and pseudo-labels are used to provide supervision on the mixed frames.
   (c) Since the mixed frames contain image patches from both source and target domains, 
   the label mixing module generates pseudo-labels based on the inputs from ground truth labels and the teacher network predictions.
   (d) The teacher network is initialized with the parameters of the student network.
   Its parameters are non-trainable and updated as the exponential moving average of the parameters of the student network.
   }}
\label{fig:overview}
\end{figure*}

\begin{figure} [htbp!]
\begin{center}
\includegraphics[width=0.99\linewidth]{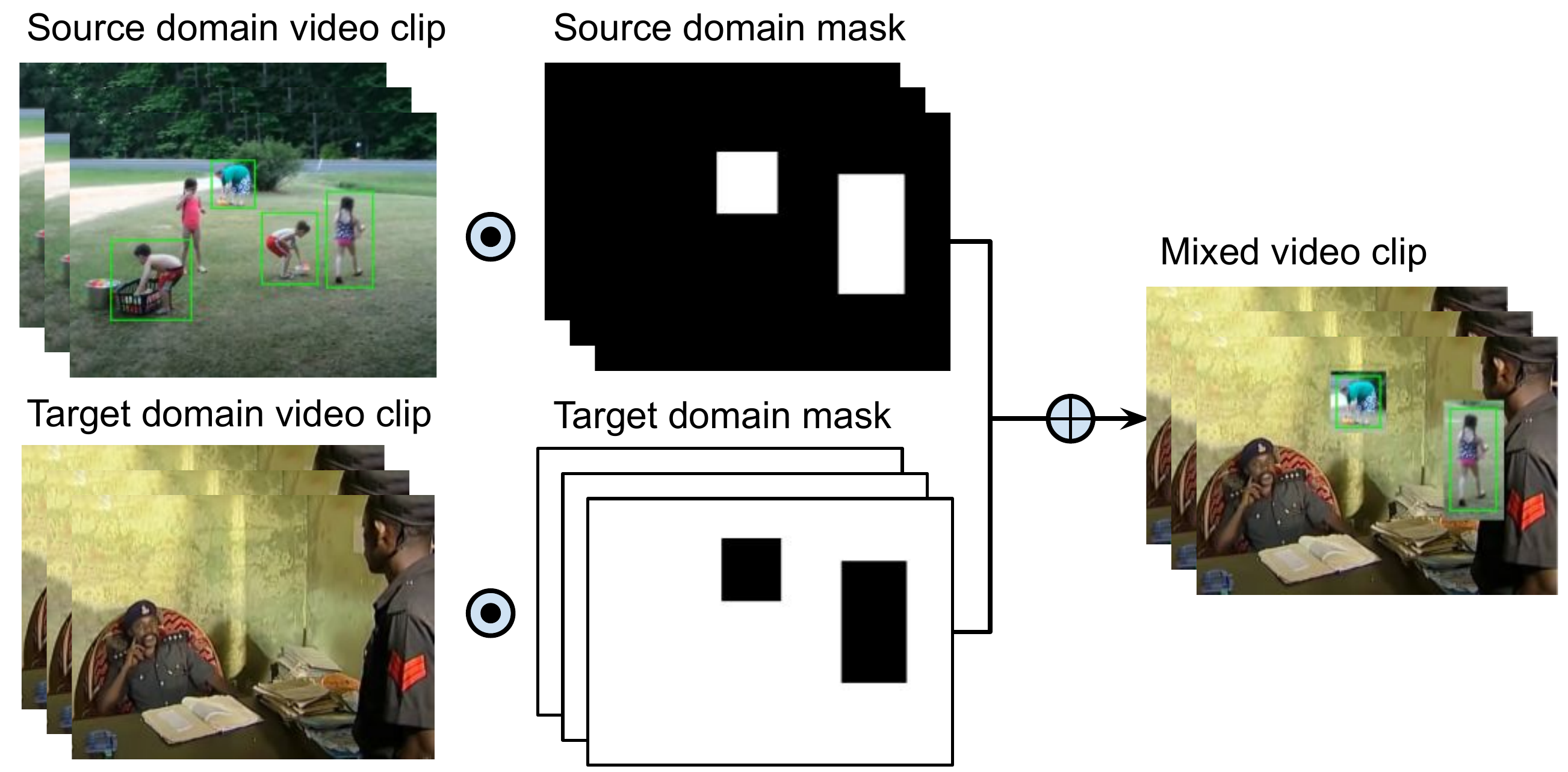}
\end{center}
   \caption{\small{The above diagram illustrates the proposed action-instance-based (AIM) cross-domain mixed sampling.}}
\label{fig:aim_mix}
\end{figure}

%% file: text/related_work.tex
\section{Related Works}
\noindent\textbf{Action Detection} is a more challenging problem~\cite{feichtenhofer2019slowfast,tuber2022zhao,singh2017online} compared to action recognition~\cite{simonyan2014twostream,carreira2017quo} problem due to the additional requirement for localisation of actions in a large spatial-temporal search space. 
Supervised action detection methods~\cite{weinzaepfel2015learning,singh2017online,kalogeiton2017action,li2020actionsas,tuber2022zhao,pan2021actor} has made large strides thanks to large scale datasets like UCF24~\cite{soomro2012ucf101}, AVA~\cite{ava2018gu} and \multisports~\cite{li2021multisports}.  Most of current approaches follow key-frame based approach popularised by \slowfast~\cite{feichtenhofer2019slowfast}. There has been more sophisticated approaches, \eg based on actor-context modelling~\cite{chen2021watch,pan2021actor}, on long-term feature banks~\cite{wu2019long,tang2020asynchronous}, and on transformer heads~\cite{tuber2022zhao,li2022mvit2}. We will make use of key-frame based \slowfast~\cite{feichtenhofer2019slowfast} network as our default action detector because of it's simplicity, competitive performance, and reproducible code base provided on \pyslowfast~\cite{fan2020pyslowfast}, which can be easily extended to include transformer architectures, such as MViTv2~\cite{li2022mvit2}. 
Apart from fully-supervised methods, there has also been works on pointly-supervised~\cite{mettes2019pointly} or semi-supervised~\cite{kumar2022end} settings. 

\noindent\textbf{Unsupervised Domain Adaptation.}
The effectiveness of UDA techniques has been studied in different vision tasks including 
image classification, object detection, semantic segmentation, action recognition and detection.
\cite{ganin2015unsupervised,haeusser2017associative,long2015learning,motiian2017unified,saenko2010adapting,sener2016learning,tzeng2014deep} propose methods to tackle DA image classification.
DA object detection is studied by \cite{saito2019strong,chen2018domain}.
Most DA semantic segmentation methods are based on
either adversarial training or self-training. 
Adversarial training follows a GAN framework \cite{ganin2016domain,goodfellow2014generative}
to aligns the source and target domains feature distributions
at 
input \cite{gong2021dlow,hoffman2018cycada}, 
output \cite{tsai2018learning,vu2019advent}, 
patch \cite{chen2018road}, 
or 
feature level \cite{hoffman2016fcns,tsai2018learning}. 
In self-training, the supervision for target domain comes from pseudo-labels \cite{lee2013pseudo}
which can be computed offline \cite{sakaridis2018model,yang2020fda,zou2018unsupervised,zou2019confidence}
or online \cite{tranheden2021dacs,wang2021domain,hoyer2022daformer}.
Consistency regularization \cite{sohn2020fixmatch,tarvainen2017mean}
or label prototypes \cite{zhang2021prototypical} 
formulated on CDMS \cite{tranheden2021dacs,zhou2021context}
or data augmentation \cite{araslanov2021self,choi2019self,melas2021pixmatch} are used to address training instabilities.
In this work, we use online self-training and consistency regularization based on CDMS.
Unlike \cite{tranheden2021dacs,wang2021domain,hoyer2022daformer,zhou2021context},
which tackle image-based  DA semantic segmentation, 
we address a video-based DA action detection.
\cite{tranheden2021dacs,wang2021domain,hoyer2022daformer,zhou2021context} use semantic class based CDMS
which show poor results in action detection.
We propose a novel action instance-based CDMS 
specifically designed to facilitate video-based action detection.

\noindent\emph{Mixed sampling.}
Within-domain and cross-domain mixing have been widely studied for image-based problems
\cite{yun2019cutmix, berthelot2019mixmatch,french2019semi, chen2021semi, tranheden2021dacs}.
Despite the effectiveness of these algorithms on
the image-based problems, mixed sampling has not been studied for video understating tasks.
We are the first to propose a novel instance-based CDMS for video action detection.


\noindent\emph{DA action recognition and detection.}
There are several methods proposed for
single-modal (RGB) \cite{chen2019temporal,choi2020shuffle,jamal2018deep,pan2020adversarial} 
or
multi-modal (RGB, flow) \cite{munro2020multi,Song_2021_CVPR,Kim_2021_ICCV} 
DA action recognition.
\cite{chen2020action,chen2020actionWACV} propose methods for DA action segmentation.
%
We found only one work \cite{agarwal2020unsupervised} that addresses DA action detection using GRL-based adversarial training.
\cite{agarwal2020unsupervised} propose two UDA benchmarks limited to sports actions.
This work has two major limitations.
Their proposed UDA setup does not address the long-tail and large variability problems (see \S \ref{sec:intro}),
and the proposed GRL-based adaptation shows a poor generalization 
in a UDA setting where 
the source domain has a long-tailed distribution, and
the class-specific actions have large variations across domains.
In contrast, our approach addresses these limitations by 
proposing a new UDA framework in which these problems are alleviated 
using an auxiliary source domain and 
a more effective instance-based CDMS and pseudo-labeling techniques.

%% file: text/method.tex
\section{Methodology}
In this section, we will introduce the proposed \gls{daaim} framework.
\gls{daaim} (Fig. \ref{fig:overview}) can be decomposed into two main steps,
namely action-instance-based CDMS (cross-domain mixed sampling) and self-training.
\begin{figure}[t!]
\begin{center}
\includegraphics[width=0.7\linewidth]{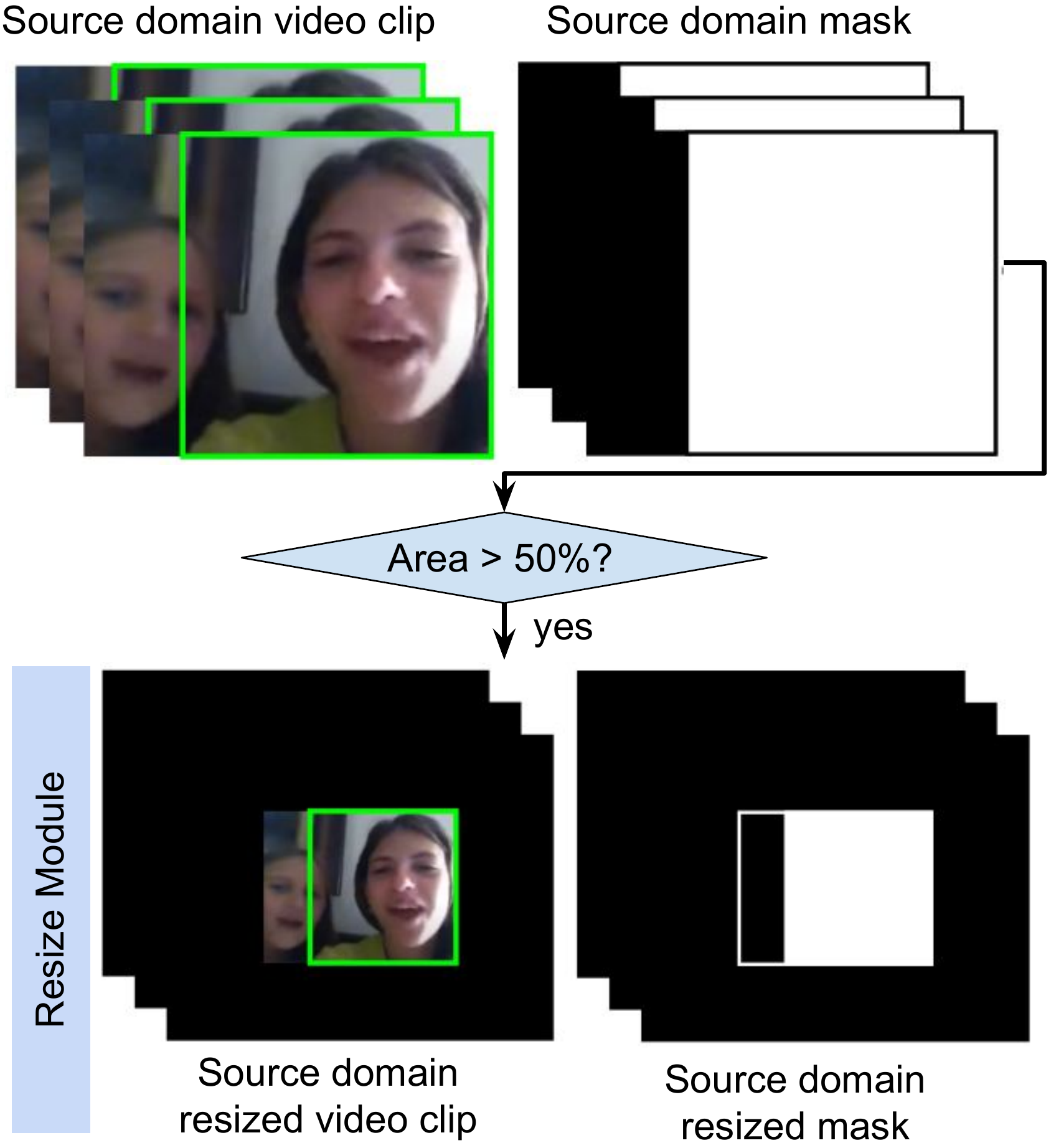}
\end{center}
   \caption{\small{Frames need to be downscaled if source domain's
   action instance area takes up more
   than half of the entire frame area.
   Bounding boxes and the mask are correspondingly
   adjusted to fit into the resized frames.
   White represents $1$ and black represents $0$.}}
\label{fig:resize}
\end{figure}
\subsection{Action-instance-based CDMS}
Fig. \ref{fig:aim_mix} illustrates the proposed Action-Instance-based cross-domain Mixed sampling (AIM).
Given video clips from the source and target domains,
and the corresponding ground truth annotations (i.e., the bounding boxes and their class labels) of the source frames,
we randomly sample half of the action instances from the source frame.
Since the bounding boxes are created only for the key-frames located in the middle of the clips,
considering fast moving actions such as running, 
we expand each bounding box by $20\%$ when creating the source domain mask.
The $3$D source domain mask $M\in \{ 0,1\}^{T\times W \times H}$ 
is constructed by replicating the $2$D mask of the key-frame $M_{k} \in \{ 0,1\}^{W \times H}$
in the temporal axis,
where $M_{k}$ is a binary matrix containing $1$ for regions where the selected source instance is present
and $0$ otherwise.
only at the places 
Our mixed video clips can be obtained through:
\begin{equation}
    x_{M} = M \odot x_{S} + (1-M) \odot x_{T} ,
\end{equation}
 where $x_{M}, x_{S}, x_{T} \in \mathbb{R}^{T\times W \times H}$ represent
 the mixed video clip, input source and target video clips respectively.

Note that often the videos from the source domain (Kinetics) 
contain action instances which take most of the image regions,
i.e., the instance bounding box has a large spatial overlap with the entire image region.
If such a video clip is used for CDMS without action instance resizing, 
it might lead to imbalance in information across domains.
That is, the mixed frames might mostly be occupied with source domain action regions,
and there would be too little target regions visible.
To address this imbalance issue, 
we propose to first resize the large action instance in the source frame and 
then paste it onto the target frame (Fig. \ref{fig:resize}).
More specifically, if the source action instance area takes up
more than half of the entire area of the mixed frame,
we will downscale the source domain frames by factor $0.5$ before mixing.
Bounding boxes and the mask are correspondingly adjusted to align with the resized video clip.
Given bounding boxes as a tuple $(x_1, y_1, x_2, y_2)$, 
where $(x_1, y_1)$ corresponds to the top left corner and $(x2, y2)$ 
corresponds to the bottom right corner, and
$H$, $W$ are the height and width of the video frames,
coordinates of bounding boxes after resizing
$(x_1^\prime, y_1^\prime, x_2^\prime, y_2^\prime)$ can be expressed as:
\begin{eqnarray}
& x_1^\prime =  [\frac{W}{4}] + [\frac{x_1}{2}], 
y_1^\prime =  [\frac{H}{4}] + [\frac{y_1}{2}]\\
& x_2^\prime =  [\frac{W}{4}] + [\frac{x_2}{2}],
y_2^\prime =  [\frac{H}{4}] + [\frac{y_2}{2}]
\label{equ:resize}
\end{eqnarray}
where $[\cdot]$ indicates the rounding function to find the nearest integer.
The empty borders after resizing are filled with 0.
Since target domain action instances might be covered by source domain action instances after mixing,
bounding boxes and labels can not be simply concatenated. 
Due to the possibility of lacking important information to identify the action,
if a bounding box from target domain overlaps with any pasted bounding boxes 
from source domain more than $40\%$ of its area,
it is discarded and not included in the loss computation.
\subsection{Self-training for UDA} 
We follow Mean Teacher~\cite{tarvainen2017mean} method for self-training.
More formally, the weights of the student network at training step $t$ is defined as 
$\theta_t$ 
and the weights of the teacher network as $\theta_t^\prime$.
At each training step $t$, weights of the teacher network $\theta_t^\prime$ are updated according to Eq.\ref{equ:meanteacher}
\begin{equation}
    \theta_t^\prime = \alpha \theta_{t-1}^\prime + (1-\alpha)\theta_t,
    \label{equ:meanteacher}
\end{equation}
where $\alpha$ is a smoothing coefficient. 
In this work, we focus on exclusive actions, 
which means those actions can not be done at the same time. 
Consequently, the problem is a single-label classification problem.
Hence, the pseudo-label of an action instance is the action class obtaining highest confidence score from the current teacher model.
\subsection{Training Optimization}
In \gls{daaim}, the student network parameters $\theta$ are trained by minimizing the following loss:
\begin{equation}
\begin{aligned}
\argmin_{\theta}\mathcal{L}(\theta)=\argmin_{\theta}\mathbb{E}\bigg[H\big(f_\theta(X_S, B_S), Y_S\big)+\\\lambda H\big(f_\theta(X_M, B_M),Y_M\big)\bigg]
\label{equ:loss}
\end{aligned}
\end{equation}
where the expectation is over batches of random variables $X_S$, $B_S$, $Y_S$, $X_M$, $B_M$ and $Y_M$. Video clips in $X_S$ are sampled uniformly from the source domain distribution, $B_S$ and $Y_S$ are the corresponding bounding boxes and labels. Furthermore, $X_M$ is the new mixed video clips, $B_M$ and $Y_M$ are mixed bounding boxes and mixed labels. As we focus on exclusive actions and formulate the problem as single-label classification, we use cross-entropy loss $H$. $\lambda$ is a hyper-parameter that decides how much the unsupervised part of the loss affects the overall training. Adapted from~\cite{tranheden2021dacs}, we use an adaptive schedule for $\lambda$, where it is the proportion of instances in the whole unlabeled instances in the mixed video clip, of which the predictions have a confidence above a certain threshold.

\subsection{DA-AIM Algorithm}  \label{sec:aim}
The overall \gls{daaim} algorithm is summarized in Alg.\ref{alg:aim}. The source-domain and target-domain datasets are referred to as $\mathcal{D}_S$ and $\mathcal{D}_T$. A batch of video clips, bounding boxes and labels, $X_S$, $B_S$ and $Y_S$, is sampled from $\mathcal{D}_S$ , and another batch of video clips, $X_T$ from $\mathcal{D}_T$. $\widehat{B}_T$ represents bounding boxes of target domain video clips estimated by a pre-trained person detector. The unlabeled video clips $X_T$ and bounding boxes $\widehat{B}_T$ are firstly fed to the teacher network $f_{\theta^\prime}$, from which pseudo-labels $\widehat{Y}_T$ are obtained. Then, the augmented video clips $X_M$ are created by mixing $X_S$ and $X_T$.The pseudo-labels $Y_M$ and bounding boxes $B_M$ are correspondingly constructed by mixing $Y_S$, $\widehat{Y}_T$ and $B_S$, $\widehat{B}_T$. Start from here, the algorithm resembles a supervised learning approach and the process is repeated for a predetermined amount of iterations $N$.

\begin{algorithm}[t!]
    \renewcommand{\algorithmicrequire}{\textbf{Input:}}
	\renewcommand{\algorithmicensure}{\textbf{Output:}}
    \caption{\gls{daaim} Algorithm}
    \label{alg:aim}
    \begin{algorithmic}[1]
    \REQUIRE $\mathcal{D}_S$, $\mathcal{D}_T$ (source and target domains), \\
    $f_{\theta^\prime}$, $f_\theta$, ${\theta^\prime}$, $\theta$ (teacher, student nets and parameters),\\
    $d_p$ (pretrained person detector).
    \ENSURE $f_\theta$ (trained student net).
    \STATE Initialize $\theta$ and $\theta^\prime$ with MiT pretrained weights. 
    \FOR {$t \gets 1,2,...,N$} 
    \STATE Randomly sample mini-batches: \\
    ($X_S$, $B_S$, $Y_S$) $\sim \mathcal{D}_S$, ($X_T$) $\sim \mathcal{D}_T$.
    \STATE Compute bounding boxes: $\widehat{B}_T\gets d_p(X_T)$.
    \STATE Compute pseudo-labels: \\
           $\widehat{Y}_T\gets\mathrm{argmax}\big( f_{\theta^\prime}(X_T, \widehat{B}_T)\big)$.
    \STATE Generate mask $M$ for mixed sampling.
    \STATE Generate the mixed video $X_M$: \\
           $X_M \gets M\odot X_S + (1-M) \odot X_T$.
    \STATE Compute pseudo-labels $Y_M$, and \\
            bounding boxes $B_M$ for $X_M$:\\ 
           $Y_M \gets CDMS(Y_S, \widehat{Y}_T)$, \\ 
           $B_M \gets CDMS(B_S, \widehat{B}_T)$. \\ 
    \STATE Forwards pass of student net $f_\theta$: \\
           $\widehat{Y}_S\gets f_\theta(X_S, B_S)$, 
           $\widehat{Y}_M\gets f_\theta(X_M, B_M)$.
     \STATE Compute cross-entropy losses:\\
            $\ell = \mathcal{L_S}(\widehat{Y}_S, Y_S) + \mathcal{L_M}(\widehat{Y}_M, Y_M)$. 
    \STATE Compute gradient $\nabla_\theta \ell$ by backpropagation.
    \STATE Optimize $\theta$ with stochastic gradient descent.
    \STATE Update ${\theta^\prime}$ using EMA (exponential moving average): \\
           $\theta_t^\prime=\alpha\theta_{t-1}^\prime+(1-\alpha)\theta_t$.
    \ENDFOR
    \RETURN $f_\theta$
    \end{algorithmic}
\end{algorithm}

%% file: text/experiments.tex
\section{Experiments and Results}

\input{tabtex/Tab_dataset_stats}
\subsection{Datasets}
We use four datasets in our experiments: AVA~\cite{ava2018gu}, \avakin~\cite{Li2020avakin}, and two in-house labelled datasets, namely InHouseDataset-1 (IhD-1) and InHouseDataset-2 (IhD-2). This section briefly introduces these datasets and describes how we use them to fit into our UDA experimental setting.

\qheading{AVA~\cite{ava2018gu}} is a dataset with atomic visual actions and consists of $430$ densely annotated $15$-minute video clips with $80$ visual actions. In total, roughly $1.62M$ action annotations are provided with the possibility that multiple annotations are made for one action instance, 
i.e., each action instance can perform multiple actions at the same time. 
We use version V2.2 of the annotation files throughout this work. In our experiments, we use AVA as one of the target domains.
For the source domain, we use \avakin. 

\qheading{\avakin~\cite{Li2020avakin}} annotates more than $200k$ videos from Kinetics-400~\cite{kay2017kinetics} dataset for action detection.
In the annotations, the AVA action classes are considered.
The ground truth bounding boxes are provided for one key-frame per $10$ seconds long video.
The main reason for using \avakin as the primary source domain is that it comes from YouTube and has high diversity compared to AVA which comes from movie clips.

\qheading{In-House Datasets.}  we build two in-house datasets using two different scenes. One dataset is recorded in a public place with varying views of the scene while actors perform one or more actions from the action list at a given time. 
The other dataset is recorded at a private facility to which access is permitted only for a limited time, and the actors are different from the former setup due to strict regulations. Going forward, former is named as In-House-Datasets-1 (\textit{\ihdone}). and later (\textit{\ihdtwo}).
These datasets contain three extra classes than \avakin or AVA datatset, namely, `carry-bag', `drop-bag', and `leave-bag-unattended'.

\input{tabtex/TAB_ablations_all}
\subsection{Dataset Sampling}
We reduce the size of the large-scale datasets due the following three reasons:
(1) action classes need to be matched to the target domain class set,
(2) for a fair comparison with smaller datasets,
(3) for minimizing time and resource consumption.
To reduce the size of the \avakin dataset, 
we set $5000$ as the maximum number of training samples for each action class.
For those action classes which have training samples less than $5000$,
all the samples from that class is considered. 
We don't reduce the size of the validation set, 
i.e., we consider all the samples from each class in the validation set.
Overall statistics of datasets used in our experiments is provided in Tab.\ref{tab:dataset_stats}. 
The table contains the statistics of each subset according to 
the number of target domain classes used in our experiments.
More details can be found in the supplementary material.
The auxiliary source domain is introduced either 
when the primary source domain does not contain one or
more target domain classes or when the primary source domain
needs help from the auxiliary source domain.

\subsection{Implementation Details}
We use \slowfast{}R50~\cite{feichtenhofer2019slowfast} as the backbone of our DA-AIM.
We implement \slowfast~\cite{feichtenhofer2019slowfast} with the help of \pyslowfast~\cite{fan2020pyslowfast}.
Since we use \avakin videos as the primary source domain, we do not want to show undue bias towards Kinetics~\cite{kay2017kinetics} dataset.
For this reason, we use the MiT pretrained weights for network weights initialization. To generate the pretrained weights, we train \slowfast{}R50 for the video classification task on MiT dataset~\cite{monfort2021multi}. 
We use \gls{map} as the metric to evaluate the performance of our proposed DA-AIM.
We use \gls{sgd} with Nesterov acceleration and a base learning rate of $1\times 10^{-2}$ for baseline experiments while $1.25\times 10^{-2}$ for others, which is then decreased using a cosine scheduler with a final learning rate equal to $1/100$ of base learning rate. Warm-up lasts $1$ epoch and starts from $1/10$ of the base learning rate. Weight decay is set to $1\times 10^{-7}$ and momentum to $0.9$. For \avakin $\to$ AVA experiments, we train on $4$ GPUs with batch size $24$ for $6$ epochs, for all other setups (\eg \avakin $\to$ \ihdtwo), we use batch size $8$ and train on $2$ GPUs for $4$ epochs.

\input{tabtex/TAB_ihd2_target}
\input{figtex/FIG_conf_mat}
\subsection{Ablation Studies}~\label{subsec:ablation}
We conduct an ablation study to investigate the efficacy of different components of our proposed \gls{daaim} method (see Tab.~\ref{tab:ablation}).
We use the following two UDA protocols for the ablation, \avakin $\to$ AVA, and \avakin $\to$ \ihdtwo.
The clear message from the above table is that we need to have all the components in place to gain substantial improvement. 

\qheading{Cross-domain instance mixing} (iMix) itself can barely promote the model to learn from the target domain, as seen in rows 3-5 of Tab.~\ref{tab:ablation}. 
Since iMix only utilizes the ground-truth labels to compute final loss, which makes the loss rely heavily on the contents from source domain while contents from target domain only have few impact. 

\qheading{Pseudo-labeling} worsens the performance on both source and target domain compared to baseline experiment without any of other \gls{da} techniques (see row 2 Tab~\ref{tab:ablation}). 
We observe that the pseudo-labels created by the teacher network tend to be biased towards easy-to-predict classes. Fig.\ref{fig:confusionmatrix} (a) illustrates the confusion matrices of real-labels vs. pseudo-labels.
In \avakin $\to$ AVA experiment pseudo-labels bias towards class \textit{sit}. 
A similar phenomenon is observed in earlier works applying pseudo-labeling to \gls{uda} for semantic segmentation tasks~\cite{zou2018unsupervised, tranheden2021dacs}.

The above-mentioned drawbacks of cross-domain instance mixing and pseudo-labeling can be redressed by integration with resizing. 
Taking pseudo-labels into consideration during loss computation pushes the network to learn domain-invariant features that apply to target domain classification as well. 
On the other hand, replacing parts of the pseudo-labels with parts of the ground-truth labels incredibly addresses the bias issue of pseudo-labels. The confusion matrices of pseudo-labels created by \gls{daaim} are present in Fig.\ref{fig:confusionmatrix} (b). 
We observe a similar trend in \avakin $\to$ \ihdtwo as well, which we can see in Tab.~\ref{tab:ablation}, the confusion matrices are presented in the supplementary material.
\qheading{Resizing} plays a crucial role in \gls{daaim}.
We verify by comparing results of cross-domain instance mixing (row 3 to row 4), and \gls{daaim} with and without resizing (row 5 and row 6) that resizing can actually enhance performance on the target domain. 

\subsection{Need for an Auxiliary Source Domain}
Here we discuss the need for an auxiliary source domain.
We need an auxiliary source domain to account for
under-represented or missing classes in the primary source domain. 
It can be observed in Tab.~\ref{tab:ihd2_target}, under-represented classes such as `take photo', `throw' and `touch' are highly benefited by the auxiliary source domain supervision.
Note the maximum performance gain (\textbf{52.48} mAP) is achieved by the model (\avakins+\ihdone), which learns meaningful representations
from both primary and auxiliary source domains for adaptation. 

\input{tabtex/TAB_sota_comparision}

\subsection{Comparison to \Sotalong}
\label{sec:sota}
We compare our \Gls{daaim} with \sotalong UDA baselines for action detection (see Table~\ref{tab:sota_comparision}).
To have a fair comparison and show the effectiveness of the proposed adaptation technique, 
 we don't consider the auxiliary source domain supervision for this set of experiments.
First, we briefly describe the baselines used for comparison. 
The the source-only baseline model is trained on the source domain and tested on the target domain.
We implement and evaluate the following three \gls{uda} baselines (for action detection) on the
proposed benchmarks (\avakin $\to$ AVA and \avakin $\to$ \ihdtwo): 
\textbf{(1)} self-supervised learning through rotation prediction (Rotation)~\cite{jing2018self},
\textbf{(2)} self-supervised learning through clip-order prediction (Clip-order)~\cite{xu2019self}, and
\textbf{(3)} adversarial training with \gls{grl}~\cite{agarwal2020unsupervised,ganin2015unsupervised}.
The oracle model is trained with full supervision (without adaptation) on the target domain dataset 
and evaluated on the validation set of the target domain.

\Gls{daaim} outperforms other \gls{da} techniques on the target domain for both \avakin $\to$ AVA and \avakin $\to$ \ihdtwo benchmarks. 
Since our evaluation benchmarks are more challenging than ~\cite{agarwal2020unsupervised}, their GRL-based approach fails to make any gains (see row second-last in Tab.~\ref{tab:sota_comparision}).
The adaption technique of the image-level approach simply fails in the challenging video-based \udalong action detection.
The same can be observed in the ablation study Section~\ref{subsec:ablation}, where simple adaption of pseudo-labeling fails.
It is important to note that our \Gls{daaim} consistently improves over other approaches, especially for the under-represented classes, e.g., `lie/sleep' and `take a photo'.
\Gls{daaim} achieves $63.65$ \gls{map} on target domain \avakin $\to$ AVA benchmark compared to $62.46$ \gls{map} of the baseline model. The improvements of average precision for class \textit{lie/sleep} and class \textit{run/jog} are more than $5\%$. Meanwhile on \avakin $\to$ \ihdtwo benchmark, \gls{daaim} increases the \gls{map} from $31.48$ of baseline experiment to $36.62$. There, the improvements of average precision for class \textit{take a photo} exceeds $10\%$. 
\begin{figure}[ht!]
    \centering
    \includegraphics[width=1.0\linewidth]{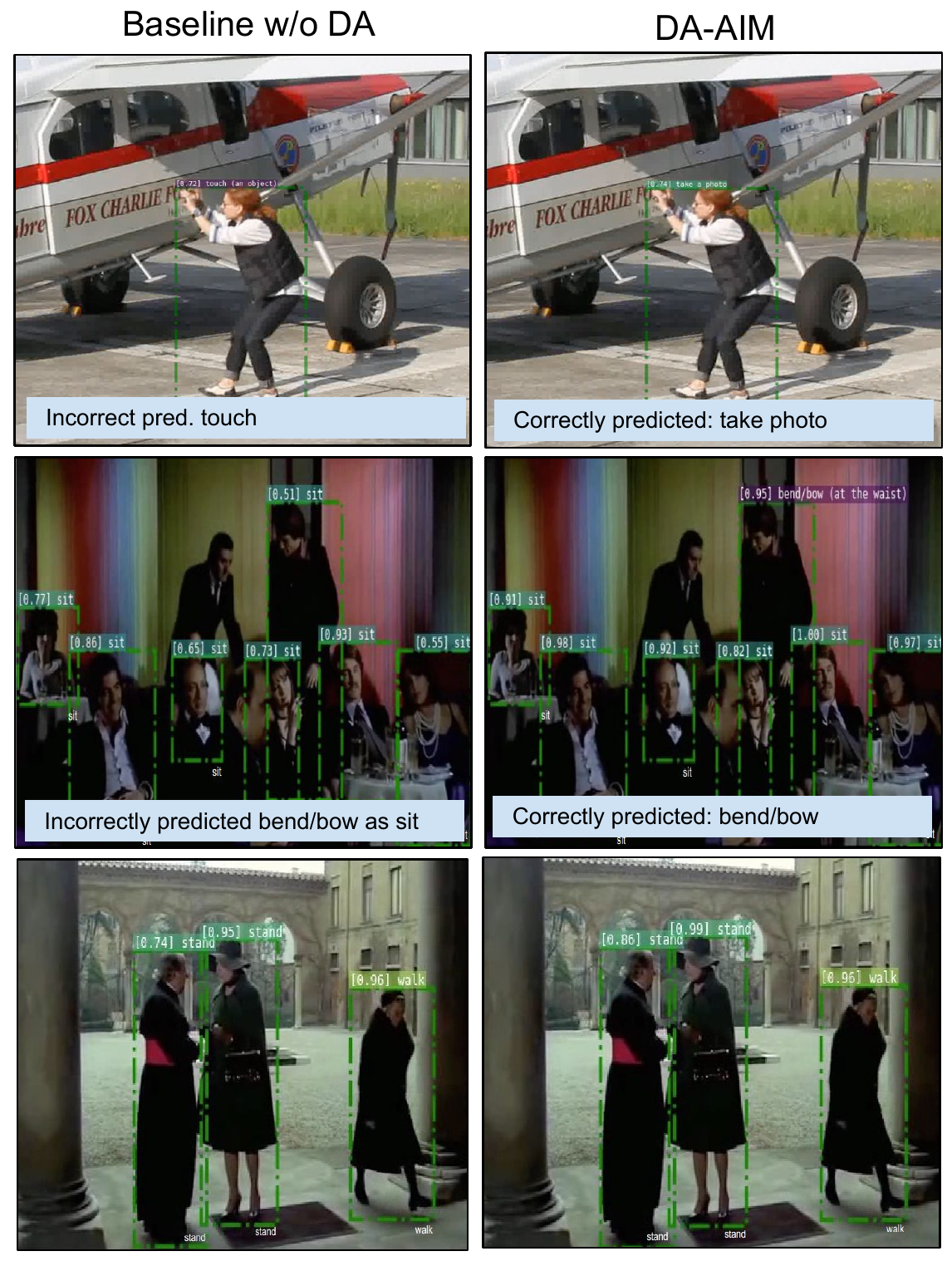}
    \caption{Visualizing the qualitative improvements made by the proposed DA-AIM method.}
    \label{fig:qualitative}
\end{figure}
It shows examples where \gls{daaim} can successfully identify difficult classes, whereas the baseline fails to do so.\\

\emph{Limitations.} We don't consider multi-label action classes, i.e., those action categories in which action instances could have more than one class label.
A rare-class sampling \cite{hoyer2022daformer} could be helpful to generate more diversified training samples for the under-represented classes.


%% file: tabtex/Tab_dataset_stats.tex

\begin{table*}[ht!]
\centering
\caption{Overall statistics of datasets used in our experiments. Each sub dataset from large-scale datasets is constructed based on number of classes in target domain and $5k$ limit on number of samples of any of the given classes.}
\setlength{\tabcolsep}{4.5pt}
\resizebox{1.00\linewidth}{!}{
\begin{tabular}{ccccccccccccc}
\toprule  
&\multicolumn{2}{c}{\textbf{AVA}}&\multicolumn{2}{c}{\textbf{\avakins}}&\multicolumn{2}{c}{\textbf{\ihdtwo}}&\multicolumn{2}{c}{\textbf{\avakins}}&\multicolumn{2}{c}{\textbf{\ihdone}}&\multicolumn{2}{c}{\textbf{\ihdtwo}}\\
&Train&Val&Train&Val&Train&Val&Train&Val&Train&Val&Train&Val\\
\midrule 
Num.of classes&\multicolumn{2}{c}{6}&\multicolumn{2}{c}{6}&\multicolumn{2}{c}{3}&\multicolumn{2}{c}{3}&\multicolumn{2}{c}{8}&\multicolumn{2}{c}{8}\\
\midrule 
Annotations&28,281&89,481&29,009&27,173&441&339&6,686&1,920&18,123&3,415&21,919&3,468\\
Unique boxes&28,281&89,481&29,009&27,173&441&339&6,686&1,920&18,114&3,415&21,843&3,442\\
Key-frames&14,248&48,741&15,453&19,205&441&339&6,115&1,779&16,881&2,695&13,974&2,753\\
Videos&235&64&15,453&19,205&12&7&6,115&1,779&28&7&34&8\\
\bottomrule 
\end{tabular}
}

\label{tab:dataset_stats}
\end{table*}


%% file: tabtex/TAB_ablations_all.tex
\begin{table*}[t]
\centering
\caption{Ablation study: impact of each operation/module introduced in our \Gls{daaim} framework for action detection.
Specifically,  impact of resizing (resize), pseudo-labeling (pLabel), and instance-mixing (iMix) modules is shown below. 
}
\setlength{\tabcolsep}{4.5pt}
\resizebox{1.00\linewidth}{!}{
\begin{tabular}{ccc ccccccc | ccc c}
\toprule  
\multicolumn{3}{c}{Operations} & \multicolumn{7}{c}{\avakin $\to$ AVA} & \multicolumn{4}{c}{\avakin $\to$ \ihdtwo}\\
resize  &  pLabel &  iMix   & bend/bow & lie/sleep & run/jog & sit & stand & walk & mAP &  touch & throw & take a photo & mAP\\
\midrule 
~ & ~ & ~  &  33.66 & 54.82 & 56.82 & \textbf{73.70} & \textbf{80.56} & \textbf{75.18} & 62.46 & 34.12 & 32.91 & 27.42 & 31.48     \\
~ & \cmarkb & ~ &  30.74 & 56.20 & 55.09 & 73.53 & \textbf{80.84} & 72.44 & 61.47 & 29.97 & 28.10 & 29.82 & 29.30  \\
~ & ~ & \cmarkb  & 33.07 & 55.87 & 60.69 & 72.51 & 79.43 & 73.05 & 62.44 & 33.00 & 29.79 & 29.26 & 30.68\\
\cmarkb & ~ & \cmarkb  & \textbf{34.65} & 56.50 & 60.19 & 70.80 & 79.17 & 74.75 & 62.68 & 32.27 & 32.48 & 30.37 & 31.71\\
~ & \cmarkb & \cmarkb  & 32.18 & 57.70 & 59.42 & \textbf{74.03} & 80.73 & 74.38 & 63.07 & 33.67 & \textbf{38.06} & 32.83 & 34.85\\
\cmarkb & \cmarkb & \cmarkb & 33.79 & \textbf{59.27} & \textbf{62.16} & 71.67 & 79.90 & \textbf{75.13} & \textbf{63.65} & \textbf{34.38} & 35.65 & \textbf{39.84} & \textbf{36.62} \\
\bottomrule 
\end{tabular}
}

\label{tab:ablation}
\end{table*}

%% file: tabtex/TAB_ihd2_target.tex
\begin{table*}[t]
\centering
\caption{Evaluation results with \ihdtwo dataset as target domain with different source domains.}
\setlength{\tabcolsep}{6pt}
\resizebox{1.00\linewidth}{!}{
\begin{tabular}{lcccccccccc}
\toprule  
 Source domain &\gls{daaim} 
 & carryBag & dropBag & leaveBag & stand& take a photo& throw& touch& walk& mAP\\
\midrule 
\ihdtwo (oracle)& \xmark &54.83 & 54.61 & 28.54 & 99.99 & 99.48 & 100.0 & 27.17 & 85.25 & 68.73\\
\midrule 
\avakins & \xmark & 37.54 & 7.36 & 1.14 & 90.72 & 96.28 & 68.40 & 2.02 & 88.18 & 48.96\\
\avakins  & \cmark & 39.97 & \textbf{9.42} & \textbf{1.26} & 86.04 & 83.71 & 76.88 & 2.08 & \textbf{89.83} & 48.65\\
\ihdone & \xmark & 18.06 & 3.12 & 0.99 & 93.17 & 98.31 & 98.62 & 4.18 & 76.04 & 49.06\\
\ihdone & \cmark & 27.75 & 7.47 & 1.16 & 94.88 & 99.26 & 97.94 & 2.70 & \textbf{81.86} & 51.63\\
\avakins+\ihdone & \xmark & 23.44 & 3.13 & 1.09 & \textbf{97.46} & \textbf{99.30} & 98.65 & 3.72 & 77.21 & 50.50\\
\avakins+\ihdone & \cmark & \textbf{42.27} & 2.77 & 1.16 & 93.45 & 98.73 & \textbf{99.01} & \textbf{7.55} & 74.89 & \textbf{52.48}\\
\bottomrule 
\end{tabular}
}
\label{tab:ihd2_target}
\end{table*}

%% file: figtex/FIG_conf_mat.tex
\begin{figure}[t]
\centering
\includegraphics[width=1.0\linewidth]{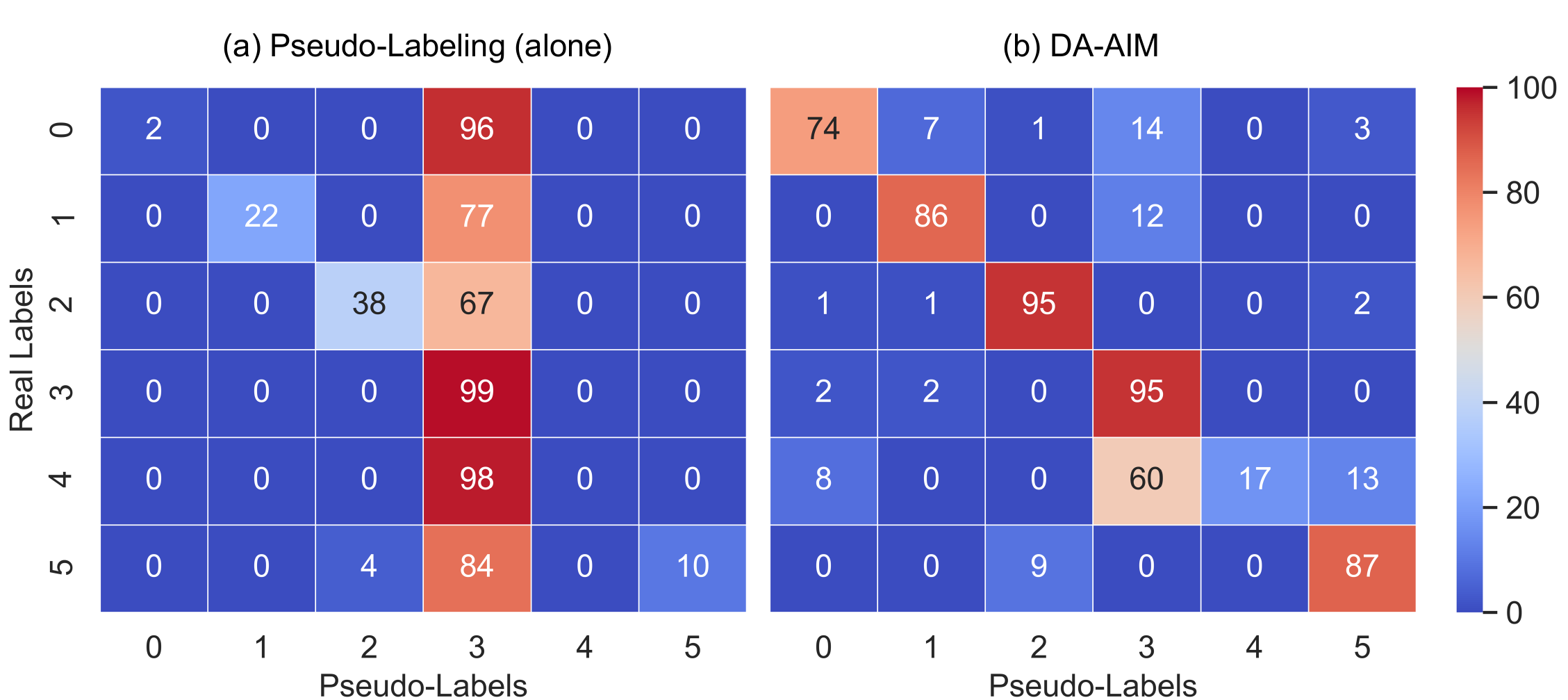}

\caption{\small{Confusion matrix of pseudo-labels at the end of training for \avakin$\to$AVA setup.
(a, left)  Pseudo-labeling alone for \uda (b, right) Pseudo-labeling within our \gls{daaim}.}}
\label{fig:confusionmatrix}
\end{figure}

%% file: tabtex/TAB_sota_comparision.tex

\begin{table*}[t]
\centering
\caption{
Comparison with state-of-the-art methods for UDA.
\Gls{daaim} is trained without the supervision of the auxiliary source domain. 
The ``source-only'' model is trained on the source domain and
evaluated on the target domain without any adaptation.
The ``oracle model'' is trained and evaluated on the target domain.
}
\setlength{\tabcolsep}{4.5pt}
\resizebox{1.00\linewidth}{!}{
\begin{tabular}{lccccccc|cccc}
\toprule  
 & \multicolumn{7}{c}{\avakin $\to$ AVA} & \multicolumn{4}{c}{\avakin $\to$ \ihdtwo} \\
Method  & bend/bow & lie/sleep & run/jog & sit & stand & walk & mAP & touch & throw & take a photo & mAP\\
\midrule
Oracle model & 36.34 & 67.49 & 57.74 & 75.61 & 84.64 & 79.26 & 66.84 & 37.91 & 51.76 & 45.38 & 45.02\\ \midrule
Source-only model & 33.66&54.82&56.82&\textbf{73.70}&\textbf{80.56}&\textbf{75.18}&62.46 &34.12&32.91&27.42&31.48\\
Rotation~\cite{jing2018self} &25.53&58.86&55.05&72.42&79.84&68.49&60.03 &30.12&34.58&25.39&30.03\\
Clip-order~\cite{xu2019self} &28.24&57.38&56.90&69.54&77.10&74.68&60.64 &28.28&32.30&29.93&30.17\\
GRL~\cite{agarwal2020unsupervised,chen2019temporal,ganin2015unsupervised} &24.99&48.41&59.89&68.68&78.79&71.38&58.69 &25.79&\textbf{39.71}&28.90&31.46 \\
\Gls{daaim} (ours) &\textbf{33.79}&\textbf{59.27}&\textbf{62.16}&71.67&79.90&75.13&\textbf{63.65}&\textbf{34.38}&35.65&\textbf{39.84}&\textbf{36.62}\\
\bottomrule 
\end{tabular}
}
\label{tab:sota_comparision}
\end{table*}

%% file: text/conclusion.tex

\section{Conclusions}
We are the first to propose a DA action detection framework based on cross-domain mixed sampling and self-training. 
We implemented and systematically analyzed the efficacy of various domain adaptation strategies including self-supervised learning, adversarial learning, self-training and naive cross-domain video mixing. 
More importantly, we proposed \gls{daaim}, a novel algorithm tailored for unsupervised domain adaptive action detection. 
\Gls{daaim} considers the inherent characteristics of action detection and mixes $3$D video clips, bounding boxes and labels (ground-truth or pseudo-labels) from source and target domain reasonably. We empirically demonstrated \gls{daaim} beat other \gls{da} techniques on two challenging benchmarks: Kinetics $\to$ AVA and Kinetics $\to$ \ihdtwo. 
Compared with baseline experiment without \gls{da} techniques, \gls{daaim} gives rise to an increase of \gls{map} by $1.2\%$ on Kinetics $\to$ AVA benchmark and $5.2\%$ on Kinetics $\to$ \ihdtwo benchmark. Average precision of class \textit{take a photo} improves over $10\%$.
In addition, we introduced the concept of auxiliary source domain. ASD domain not only help to improve the performance of \gls{daaim} on classes that are missing in primary source domain but also help other under-represented classes in long-tailed primary source domain.

\textbf{Acknowledgments.} The authors gratefully acknowledge the support by Armasuisse.

%% file: text/sup-mat/dataset_creation.tex
In this document, we provide supplementary materials for our main paper submission. 

\section{Proposed UDA Protocols}
Unlike domain-adaptive (DA) semantic segmentation \cite{ganin2016domain,hoffman2018cycada,hoffman2016fcns}, 
for DA action detection, there is no standard UDA training/validation protocol available \cite{agarwal2020unsupervised}.
The main reason is the lack of suitable pairs of source and target domains (datasets) which have common action classes.
Agarwal \etal \cite{agarwal2020unsupervised} proposed two UDA protocols (UCF-Sports $\to$ UCF-101, JHMDB $\to$ UCF-101)
which are limited to only three/four sports-related actions (e.g., ``diving'', ``golf-swing'', ``horse-riding'', ``skate-boarding'').
Besides, UCF-Sports \cite{agarwal2020unsupervised}, UCF-101 \cite{agarwal2020unsupervised} and JHMDB \cite{agarwal2020unsupervised} datasets are quite outdated.
In this work, firstly we propose a new UDA protocol \emph{\avakin $\to$ AVA}
which uses two recent action detection datasets, \avakin \cite{Li2020avakin} and AVA \cite{ava2018gu}.
These new datasets are more large-scale and diversified as compared to UCF-Sports, UCF-101 and JHMDB,
and thus, would be useful to learn better generalizable representation useful for adaptation task.
Moreover, we propose two new action detection datasets, \ihdone and \ihdtwo which allow us two
explore several new UDA protocols as shown in Table \ref{tab:uda_proto}.
Please note, our proposed UDA framework allows us to consider a wider range of action classes 
by leveraging an auxiliary source domain and it is not limited to a certain kind of actions.
Our new datasets could be also useful for suspicious action detection.
For instance, we introduce a set of new action classes (\emph{carryBag}, \emph{dropBag} and \emph{leaveBag}), where
\emph{carryBag}: ``person carrying a bag'', \emph{dropBag}: ``person dropping a bag on the floor'', 
\emph{leaveBag}: ``person leaving a bag unattended''.
These actions quite often performed in a sequnetail manner.
Among these three actions, two of them are regular actions, i.e., \emph{carryBag} and \emph{dropBag}.
But, \emph{leaveBag} might be a suspicious one.

\emph{We hope, our proposed UDA protocols facilitate exploring new research directions in domain-adaptive action detection}.
\begin{table*}[h]
\centering
\caption{
UDA protocols used in this work for training and evaluation of the proposed domain-adaptive action detection model.
ASD: auxiliary source domain, MS: main paper submission. The ``Table'' column shows the table number in which the experimental results are reported for a 
particular UDA protocol. The ``+'' symbol denotes the sample mixing step between the primary and auxiliary source domains.
For training, labeled samples from source domain (positioned at the left side of the arrow $\to$),
and unlabeled samples from the target domain (positioned at the right side of the arrow $\to$)
are used. Validation is always done on the target domain validation set.
}
\setlength{\tabcolsep}{6pt}
\resizebox{0.9\linewidth}{!}{
\begin{tabular}{lclcc}
\toprule  
UDA Protocol & classes & labels & ASD & Table \\
\midrule 
\avakin $\to$ AVA & 6 &  bend/bow, lie/sleep, run/jog, sit \& stand, walk & - & MS:2 \\
\avakin $\to$ \ihdtwo & 3 & touch, throw, take a photo & - & MS:2 \\
\avakin $\to$ \ihdtwo & 8 & carryBag, dropBag, leaveBag, stand, take a photo,  throw, touch, walk & - & MS:3 \\
\ihdone $\to$ \ihdtwo & 8 & carryBag, dropBag, leaveBag, stand, take a photo,  throw, touch, walk & - & MS:3 \\
\avakins+\ihdone $\to$ \ihdtwo & 8 & carryBag, dropBag, leaveBag, stand, take a photo,  throw, touch, walk & \ihdone & MS:3 \\
\bottomrule 
\end{tabular}
}
\label{tab:uda_proto}
\end{table*}
\subsection{Dataset Creation} \label{subsec:dataset_creation}
\begin{figure}[t!]
\begin{center}
   \includegraphics[scale=0.6]{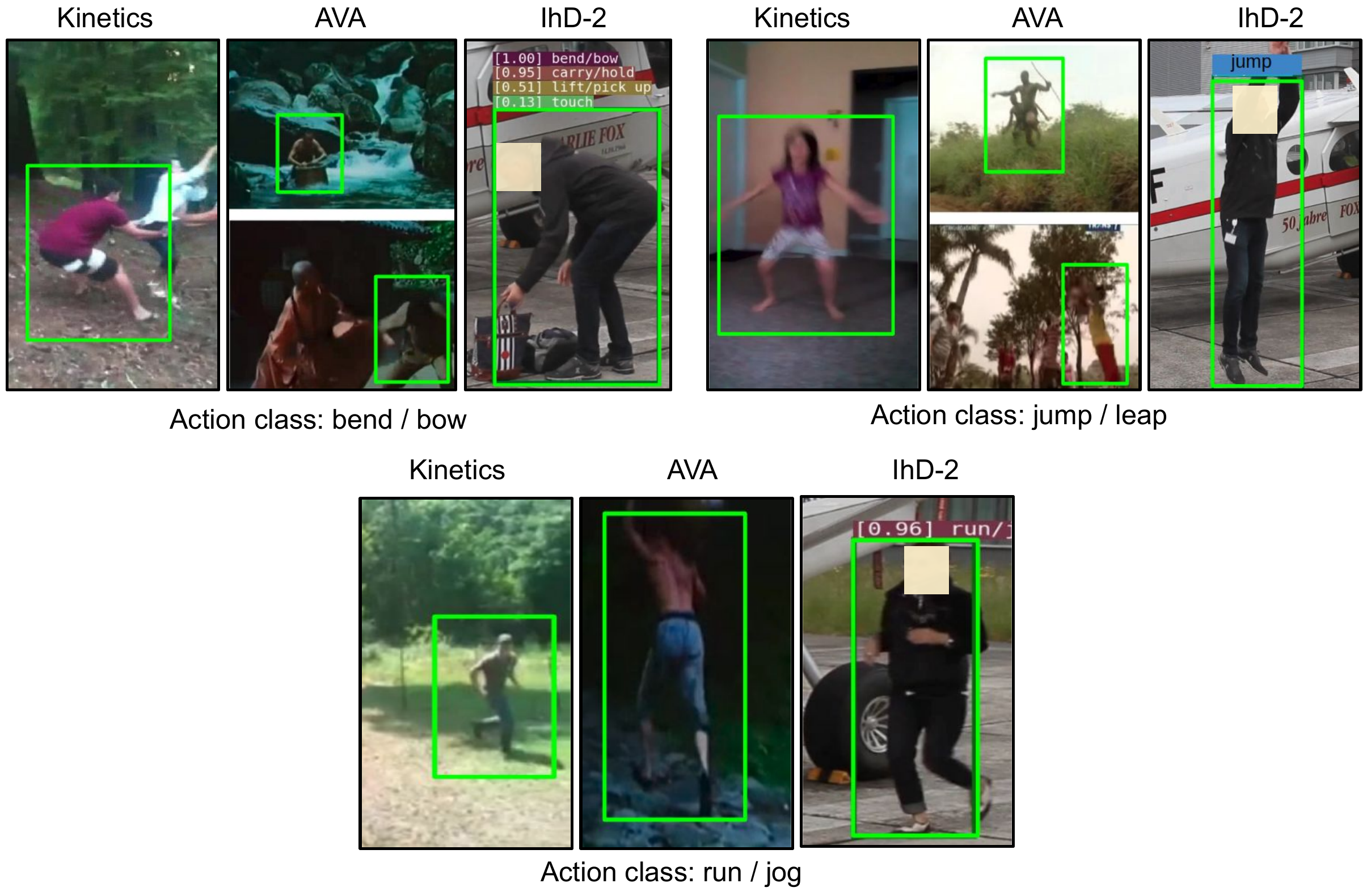}
\end{center}
\caption{\small{Illustrating large variability of action instances of the same class across three different domains (or datasets): 
Kinetics (public), AVA (public) and IhD-2 (private).
For maintaining anonymity, we cover the subject faces.
}}
\label{fig:large_variability}
\end{figure}

A real-world setting highly influences our dataset creation process 
in which access to the target domain data is limited.
That is, the target domain videos were captured at a private facility to which access is permitted only for a limited time resulting in a small number of actors and videos.
Furthermore, the action categories we are interested in detecting in the target domain are heavily under-represented in the source domain due to the long-tail distribution problem.
Another thing to note is that there is a large variability in actions (belonging to the same action categories) across domains (see Fig. \ref{fig:large_variability}). 
\begin{figure}[h]
\begin{center}
   \includegraphics[scale=0.9]{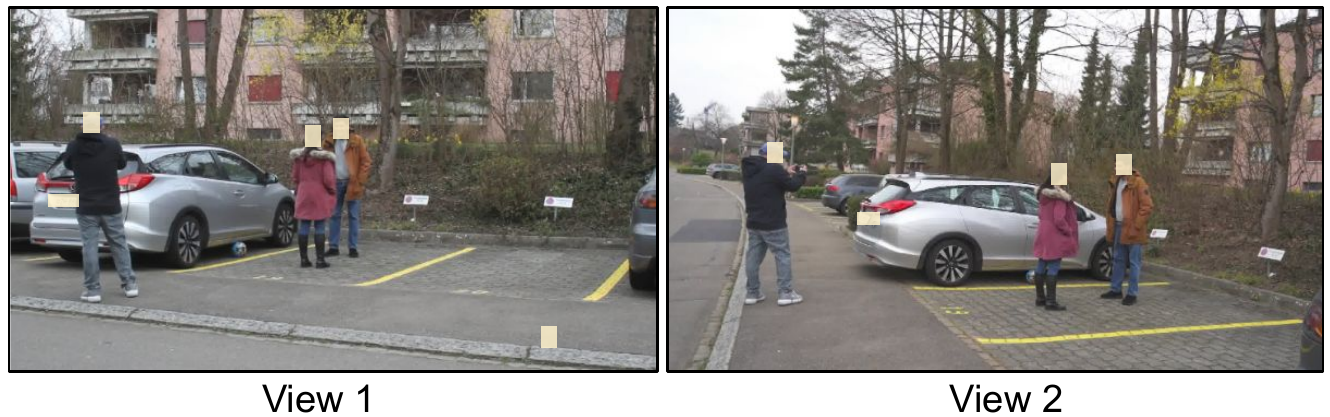}
\end{center}
\caption{\small{Action scenes are recorded using two camera from two different viewing angles.
Two sample frames captured from two different views of the same action scene are shown here.
Samples belong to the IhD-1 private dataset proposed in this work.
For maintaining anonymity, we cover the subject faces.
}}
\label{fig:dif_vews}
\end{figure}
\begin{figure*}[h]
\begin{center}
   \includegraphics[scale=0.8]{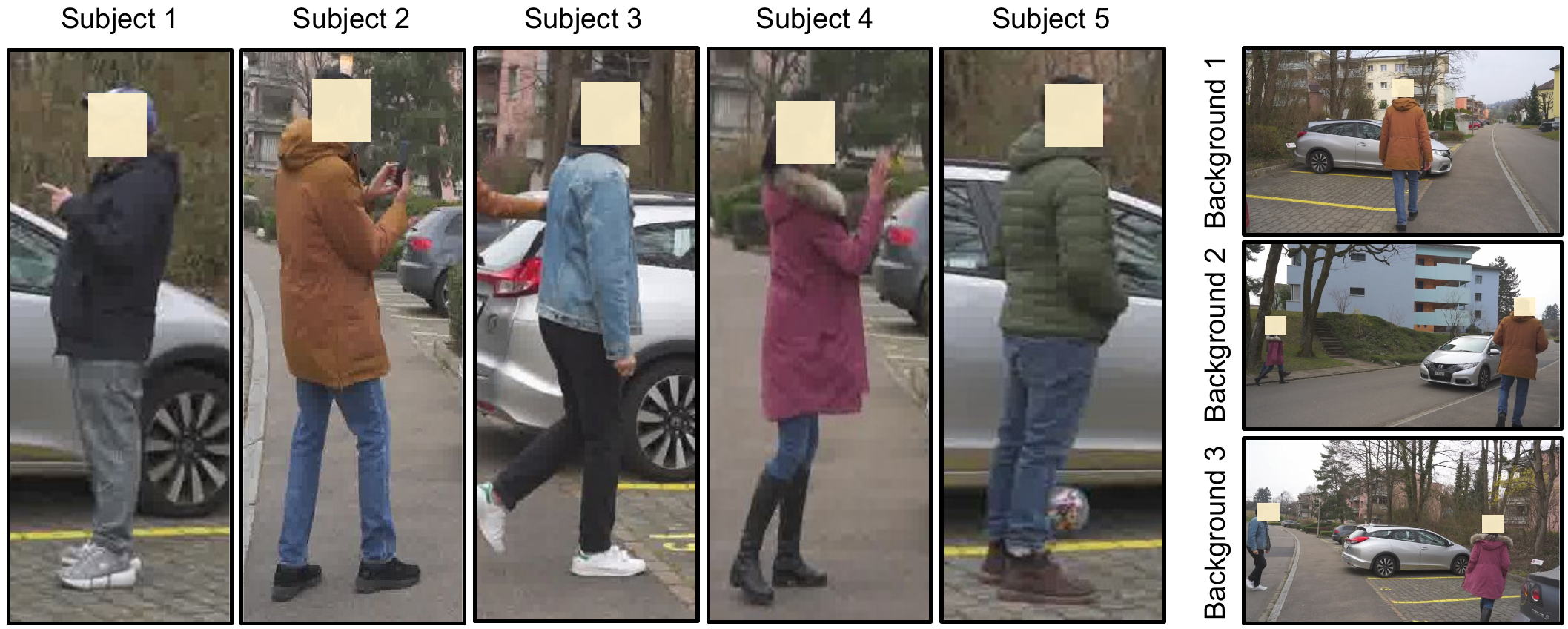}
\end{center}
\caption{\small{Five subjects and three backgrounds are used in IhD-1 dataset.
For maintaining anonymity, we cover the subject faces.
}}
\label{fig:actors_backgrounds}
\end{figure*}
\begin{figure*}[t!]
\begin{center}
   \includegraphics[scale=0.8]{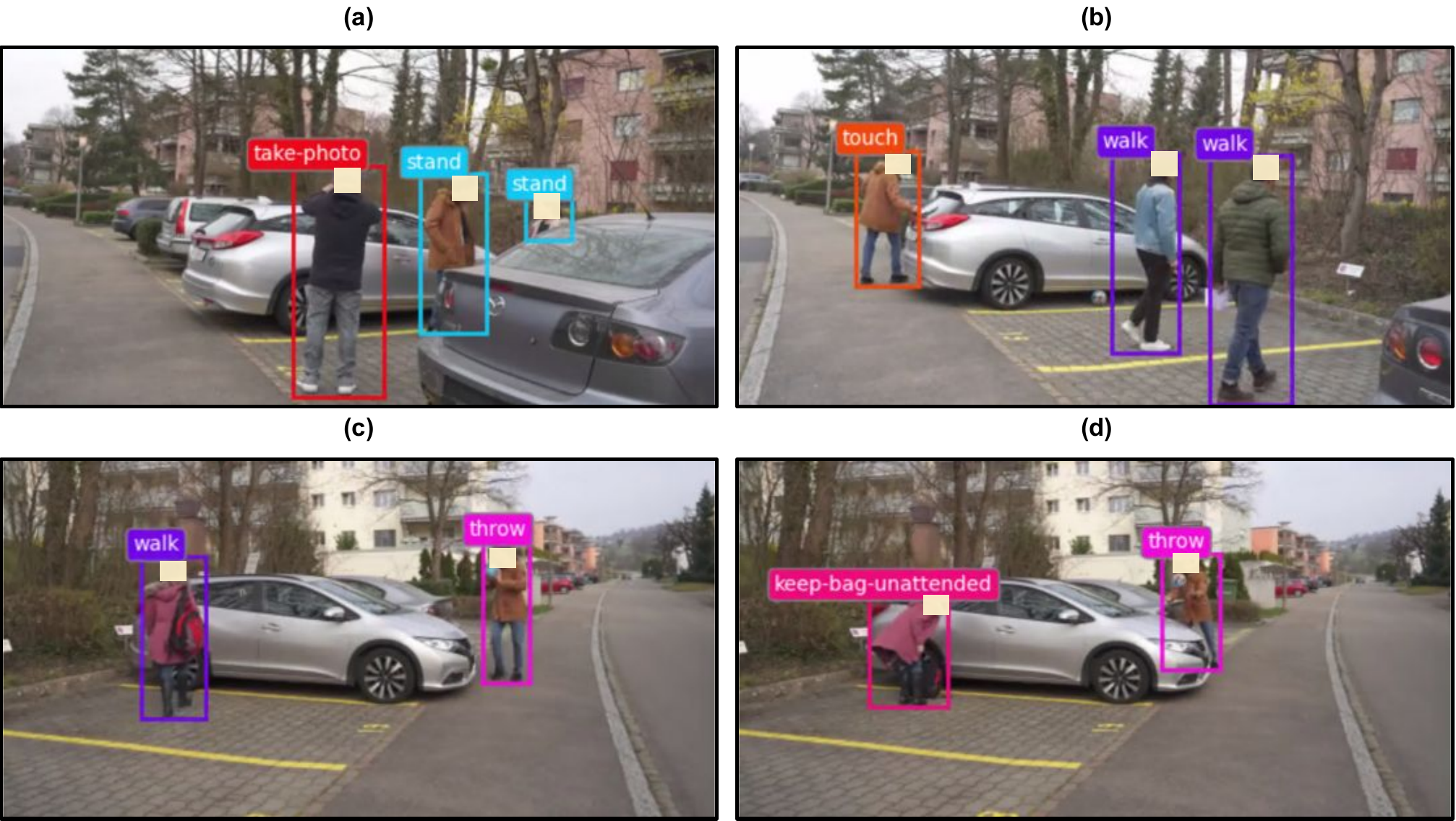}
\end{center}
\caption{\small{Sample frames from IhD-1 dataset demonstrating co-occurring action instances.
(a) ``take-photo'', ``stand''; 
(b) ``touch'', ``walk'',
(c) ``walk'', ``throw'',
and
(d) ``keep-bag-unattended'', ``throw''.
Bounding boxes depict ground truth annotations.
Each unique color denote an action class.
For maintaining anonymity, we cover the subject faces and any relevant information.
}}
\label{fig:co-occurring-actions}
\end{figure*}
\begin{figure*}[h]
\begin{center}
   \includegraphics[scale=0.3]{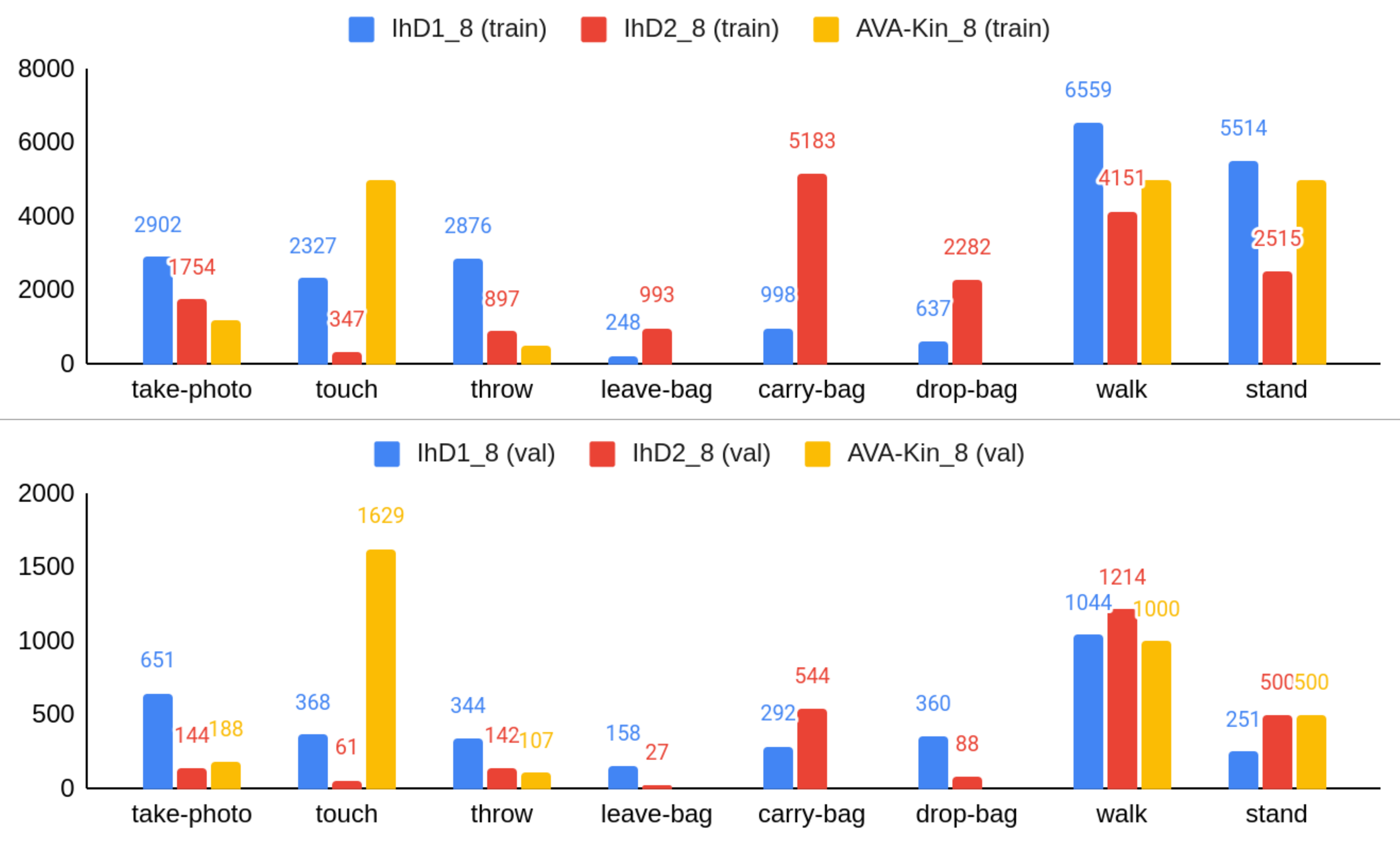}
\end{center}
\caption{\small{
Samples per class for datasets \avakin (5 classes) and \ihdone (8 classes), \ihdtwo (8 classes) for respective train and validation sets.
}}
\label{fig:plot_ihd1_8_ihd2_8_ava_kin_5}
\end{figure*}
\begin{figure*}[h]
\begin{center}
   \includegraphics[scale=0.3]{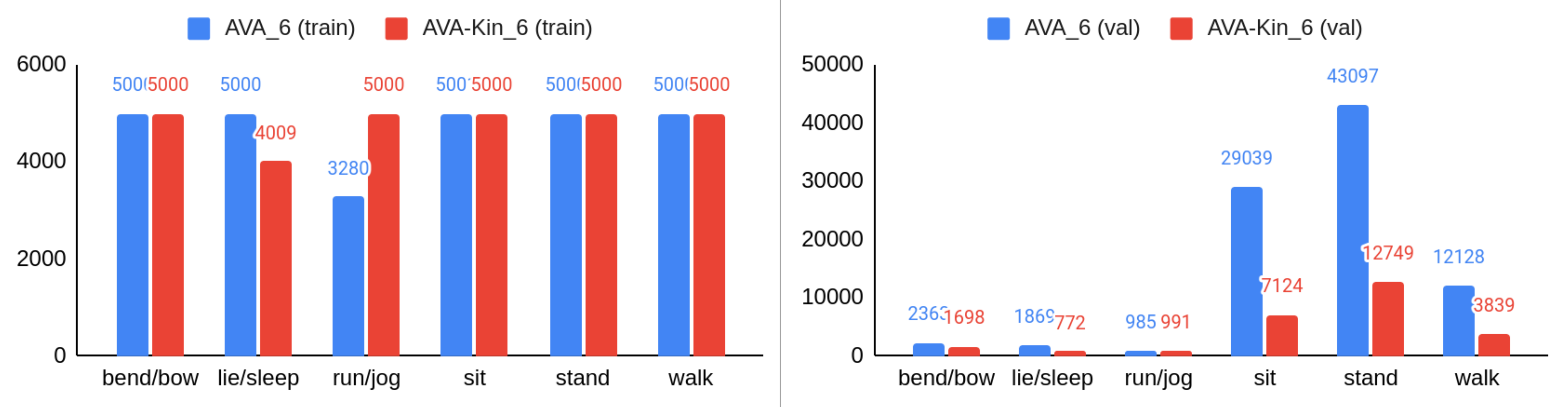}
\end{center}
\caption{\small{
Samples per class for datasets \avakin (6 classes) and AVA (6 classes) for respective train and validation sets.
}}
\label{fig:plot_ava_kin_6_ava_6}
\end{figure*}
\begin{figure*}[h]
\begin{center}
   \includegraphics[scale=0.3]{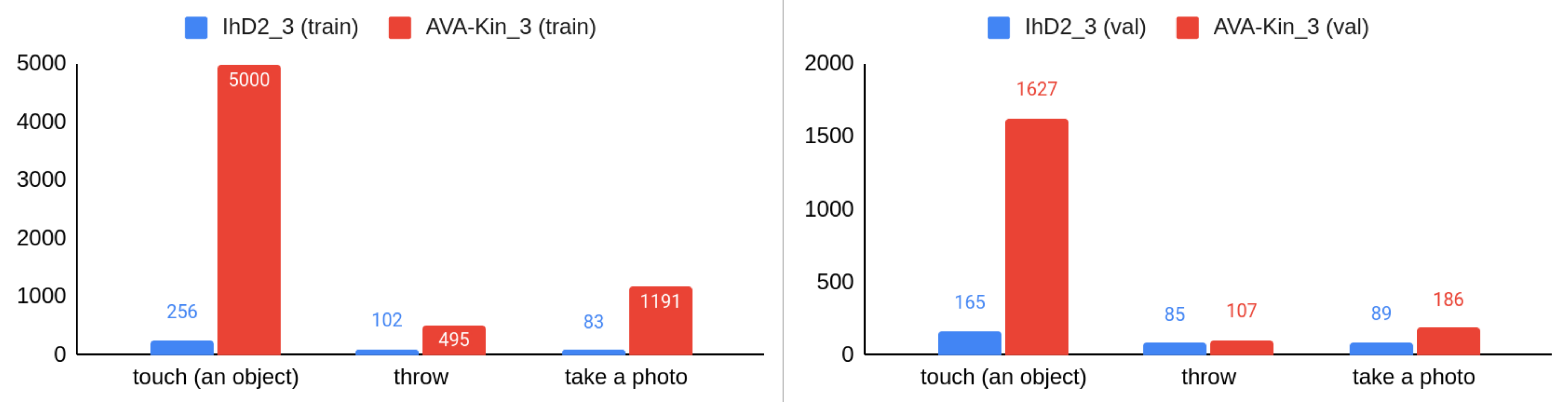}
\end{center}
\caption{\small{
Samples per class for datasets \avakin (3 classes) and \ihdtwo (3 classes) for respective train and validation sets.
}}
\label{fig:plot_ava_kin_3_ihd2_3}
\end{figure*}
\begin{figure*}[h]
\begin{center}
   \includegraphics[scale=0.6]{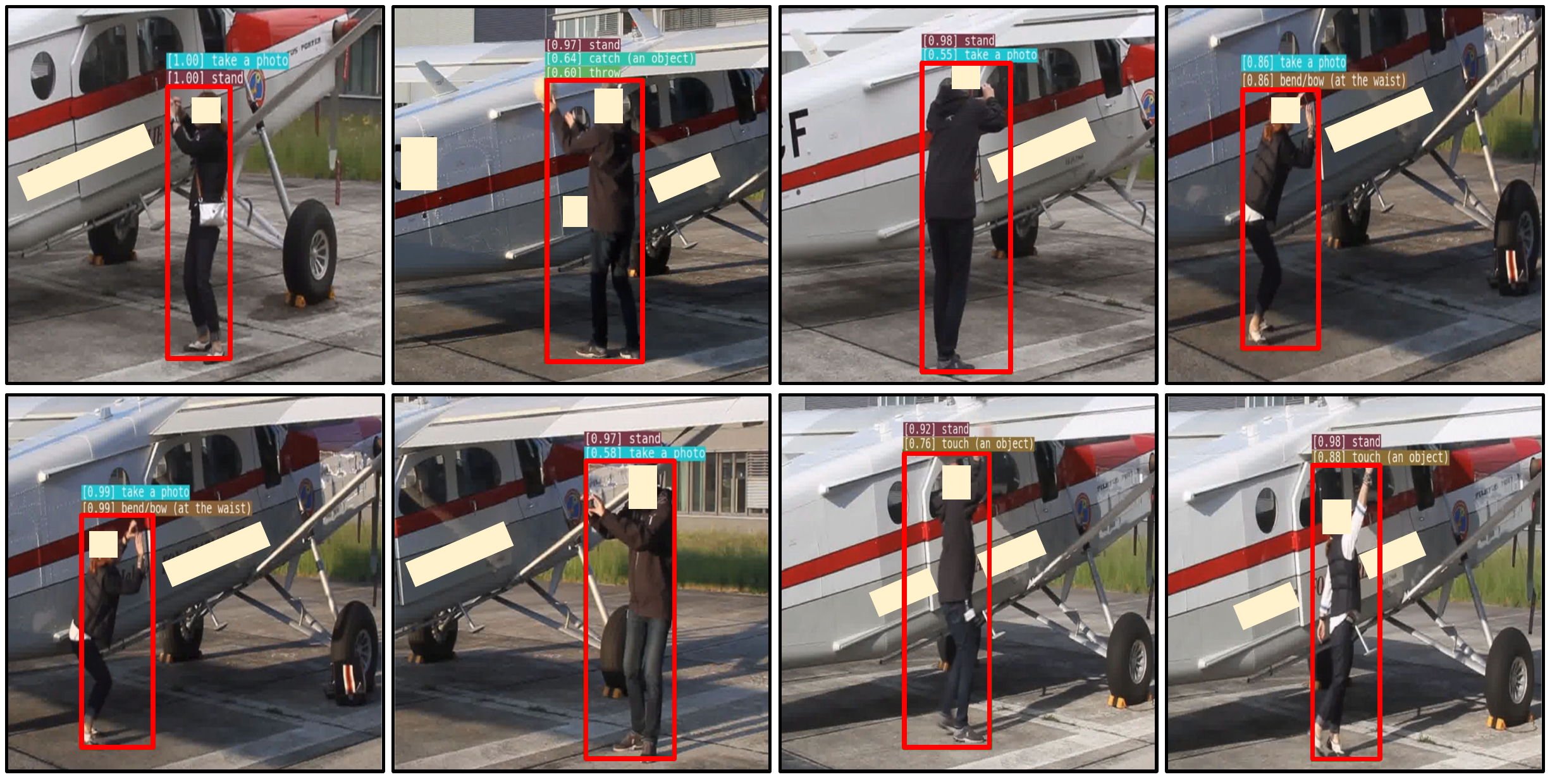}
\end{center}
\caption{\small{
Qualitative DA action detection results of our proposed model trained on UDA protocol
\avakins+\ihdone $\to$ \ihdtwo. 
Sample detection results are shown on the validation set of \ihdtwo dataset.
Our DA-AIM can successfully detect action classes such as ``take-a-photo'', ``touch'', ``throw'' and  ``stand''.
For maintaining anonymity, we cover the subject faces and any relevant information.
}}
\label{fig:qualitatives-ihd2}
\end{figure*}
\begin{figure*}[h]
\begin{center}
   \includegraphics[scale=0.6]{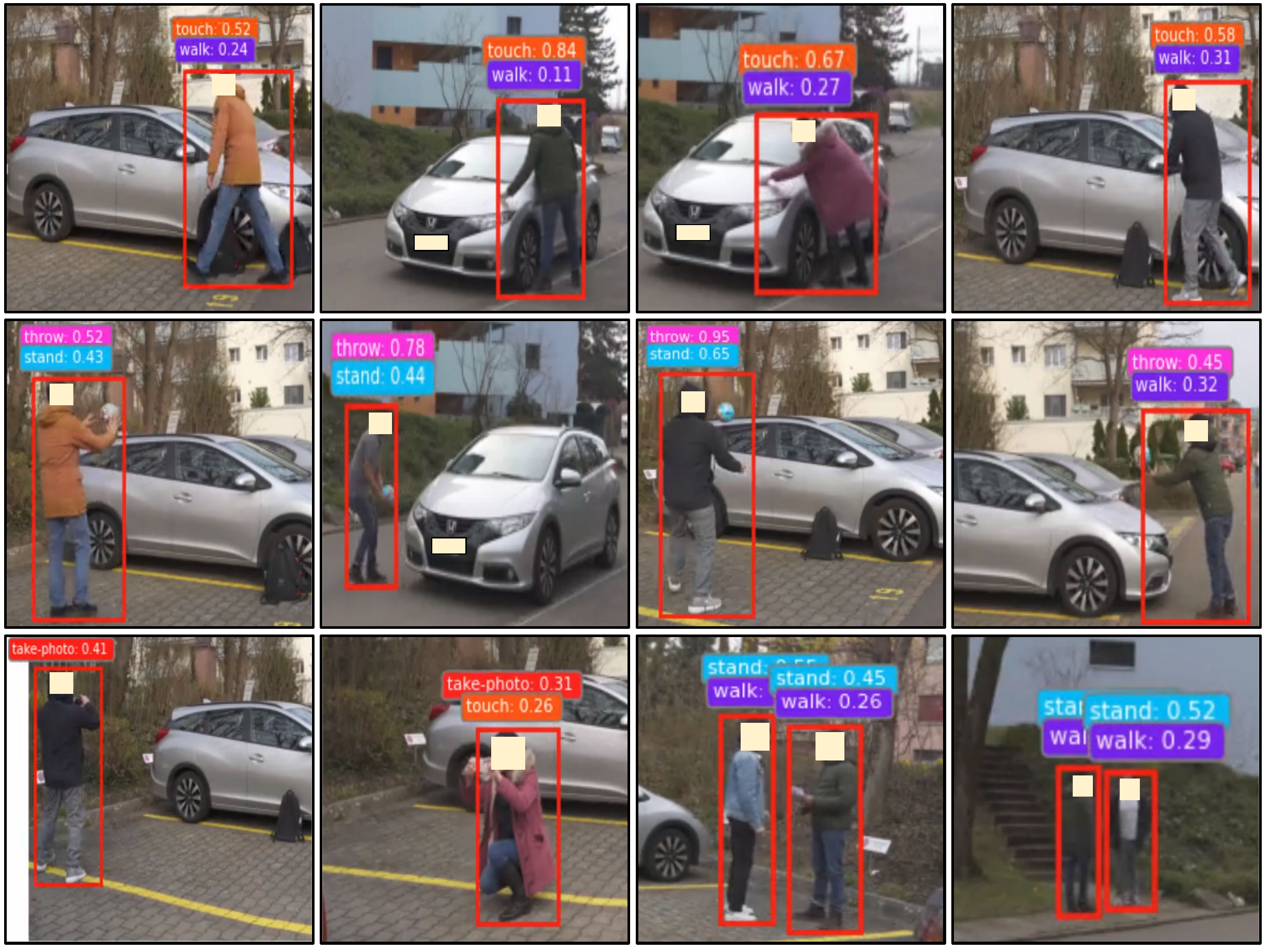}
\end{center}
\caption{\small{
Qualitative DA action detection results of our proposed model trained on UDA protocol
\avakins+\ihdtwo $\to$ \ihdone. 
Sample detection results are shown on the validation set of \ihdone dataset.
Our DA-AIM can successfully detect action classes such as ``touch'', ``throw'', ``take-a-photo'',  ``stand'' and ``walk''.
For maintaining anonymity, we cover the subject faces and any relevant information.
}}
\label{fig:qualitatives-ihd1}
\end{figure*}

To address these two problems, we propose a new UDA framework that facilities the model training by providing ground-truth supervision from an auxiliary source domain (ASD).
ASD alleviates the long-tail distribution by injecting more training samples
of those classes which are under-represented in the source domain.
The videos of the ASD are captured in a public place 
that is easily accessible without restriction, allowing us to capture more videos with multiple actors.
Since we have access to the unlabeled training data of the target domain
and thus, we already know:
what are the actions present there,
roughly their appearance and motion patterns.
We make use of these priors to generate the videos of the ASD.
More specifically, 
we record videos of those action classes which are present in the target domain,
and try to resemble (as much as possible) the action scenes of the target domain 
while recording the videos for ASD.
To this end, we create two \textbf{i}n-\textbf{h}ouse action detection \textbf{d}atasets (IhD):
(1) IhD-1 and (2) IhD-2.

\qheading{IhD-1.} The videos of IhD-1 are recorded in a public place which is easily accessible without any restrictions.
Same action scene is recorded using two cameras to get two different view of the same scene.
Fig. \ref{fig:dif_vews} illustrates two different views of the same scene. 
It facilitates adaptation across scenes within the same domain. 
We keep this setting for future exploration and use only videos from one view in this work.
To induce diversity, we use three different backgrounds 
and five different subjects (actors) (see Fig. \ref{fig:actors_backgrounds}) .
Moreover, actors change their clothes alternatively in-between two action scenes to bring variations in the appearance.
We generate action videos of co-occurring action instances (of same or different action classes) to simulate real-world scenarios.
Fig. \ref{fig:co-occurring-actions} shows some examples of co-occurring action instances.

\qheading{IhD-2.} 
The videos of \ihdtwo were recorded in a private area and access to the place is limited.
That is multiple entries are not allowed and there is a strict time limit to capture some 
sample videos.
Also, due to security reasons, only two subjects or actions are allowed to perform different actions.
Due to these limitations, \ihdtwo (target domain) has very view train and validation samples with 
less diversity in the training set.

\qheading{Video Annotation Process.}
We use the VoTT (Visual Object Tagging Tool) \cite{microsoft_vott} to annotate the videos of IhD-1 and IhD-2.
Two human annotators were assigned for the annotation task.
First, the videos are loaded to VoTT and key-frames are selected for annotation.
For each video, key-frames are selected at a frame rate of 4 FPS.
For instance, video with 30 seconds duration would have 120 key frames.
For each key-frame, bounding box annotations and their corresponding class labels are provided.
For generating dense frame-level annotation, we guide the annotation process by a 
YOLO-V5\cite{glenn_jocher_2021_5563715} person detector. 
More specifically, we propagate the key-frame ground truth bounding boxes in time 
for the regular frames by using a simple tracking algortihm.
The tracking algorithm first localize the action instances in the first key-frame using the VoTT ground truth boxes.
Next, for each regular frame where there is no ground truth box available,
it picks the YOLO-V5 bounding boxes and match them with the previous key-frame's ground truth boxes.
The set of best matched boxes are used as ground truth boxes for the current frame.
The matching is done based on the intersection-over-union (IoU) scores among the 
ground truth and YOLO-V5 boxes.
For a sanity check of the YOLO-V5 person detector,
we run inference on the AVA-Kinetics validation set
and found that the recall to be very high.
For both IhD-1 and IhD-2 videos are recorded with a frame rate of 30 FPS.
The spatial dimension of the video frames is $920 \times 1080$ pixels.

\qheading{Dataset Statistics.}
We use the following datasets in this work:
(1) AVA-Kinetics (6 classes),
(2) AVA (6 classes),
(3) AVA-Kinetics (3 classes),
(4) IhD-2 (3 classes),
(5) AVA-Kinetics (5 classes),
(6) IhD-1 (8 classes),  and
(7) IhD-2 (8 classes).
AVA-Kinetics-6, -3, and -5 are the subsets of the original AVA-Kinetics dataset
and thus they belong to the same domain.
AVA-6 is a subset of the original AVA dataset.
IhD-2-3 is the subset of the proposed IhD-2-8 dataset.
Fig \ref{fig:plot_ihd1_8_ihd2_8_ava_kin_5}, 
\ref{fig:plot_ava_kin_6_ava_6}, and 
\ref{fig:plot_ava_kin_3_ihd2_3}
show the bar plots depicting the per-class sample distribution for the training and validation sets for these datasets.
\\
Please note in Fig. \ref{fig:plot_ihd1_8_ihd2_8_ava_kin_5},
the number of training samples for classes ``take-photo''  and ``throw''
are very less in the source domain (AVA-Kinetics) and restricted target domain (\ihdtwo).
Our proposed ASD helps alleviate these data imbalance issue 
by injecting labeled training samples for these classes.
Although, there are sufficient number of training samples available for 
class ``touch'' in the source domain, but due to large variability between the source and target domain's 
data distribution, the adaptation from the source to target domain is ineffective.
Our ASD address this domain shift by generating more training samples of ``touch'' action
in a setting where the action scenes resembles to the target domain's scenes.
Furthermore, for the missing actions such as ``carry-bag'', ``drop-bag'' and ``leave-bag'',
our ASD provides more lableled traing samples to provide better supervision to the model.
One important thing to note that, although the plot shows more number of training samples for 
``bag'' related actions in the target domain (\ihdtwo), but
these samples are homogeneous (or less diversified).
That is, the action scenes in these video frames have limited 
number of actors, backgrounds due to the fact that the target domain has very limited access.
One the other hand, the ASD's samples are more diversified with more number of actors, backgrounds.

\section{Pretrained Weights for UDA}
We use \slowfast~\cite{feichtenhofer2019slowfast} as our backbone network.
There are pretrained weights publicly avaibale for \slowfast at \pyslowfast~\cite{fan2020pyslowfast}.
These pretrained weights are generated by training the \slowfast network on the \avakin dataset.
Since, we use \avakin videos as primary source domain, 
we do not want to show undue bias towards Kinetics~\cite{kay2017kinetics} dataset, 
we pretrain \slowfast{}R50 for video classification task on MiT dataset~\cite{monfort2021multi}. 
We will make the pretrained weights publicly available upon the acceptance of paper.
MiT dataset~\cite{monfort2021multi}  consits of 305 action/event classes.
It has 727,305 training videos and 30,500 testing videos.
We train the \slowfast network on MiT using 8 GPUs (GeForce RTX 2080 TI) for 10 days.


\section{Additional Quantitative Results}
\begin{table}[t]
\centering
\caption{
Comparison of the source-only model performance with the proposed DA-AIM.
The source-only model is trained on the \avakin dataset.
The DA-AIM is trained following the proposed UDA protocol \avakin $\to$ \ihdone.
Both the models are evaluated on the validation set of \ihdone.
Note that the proposed UDA protocol helps improving the action recognition performance for 
certain classes (``throw'', ``touch'' and ``walk'') on the unseen target domain samples.
}
\setlength{\tabcolsep}{4.5pt}
\resizebox{0.55\linewidth}{!}{
\begin{tabular}{lcccccc}
\toprule  
Models          & stand & take-photo & throw & touch & walk  & mAP \\
\midrule
Source-only     & \textbf{41.0} & \textbf{94.3}  & 46.6           & 15.3           & 60.9            & 51.6 \\
DA-AIM          & 22.2              & 94.0       & \textbf{47.4}  & \textbf{33.4}  & \textbf{68.6}   & \textbf{53.1} \\
\bottomrule 
\end{tabular}
}
\label{tab:ava_to_ihd1}
\end{table}
\subsection{Effectiveness of the proposed UDA protocol}
In this section, we discuss the benefits of the proposed UDA protocol \avakin $\to$ \ihdone.
For this UDA protocol,  we have created a new action detection dataset \ihdone.
Please refer to \ref{subsec:dataset_creation} for information on dataset creation.
In Tab. \ref{tab:ava_to_ihd1}, we report the results of the source-only and DA-AIM models.
The DA-AIM is trained following the proposed UDA protocol \avakin $\to$ \ihdone.
Note that the proposed UDA protocol helps improving the action recognition performance for 
certain classes (``throw'', ``touch'' and ``walk'') on the unseen target domain samples.

\subsection{DA-AIM improves pseudo-labels}
In this section, confusion matrices (real-labels vs. pseudo-labels)
of different UDA models are presented.
Fig. \ref{fig:avakin2ihd2_cm} compares the confusion matrices
of two different models.
The models are trained following the \avakin $\to$ \ihdtwo UDA protocol on three classes.
The pseudo-label-only model (Fig. \ref{fig:avakin2ihd2_cm}\textbf{a}) is trained using only pseudo-labels without the DA-AIM.
The DA-AIM model (Fig. \ref{fig:avakin2ihd2_cm}\textbf{b}) is trained following the proposed approach.
Note that the bias from the pseudo-labels in the  pseudo-label-only model is
is rectified by the DA-AIM approach.
\input{figures/sup-figtex/FIG_conf_mat_avakin2ihd2_3}

However, sometimes due to the presence of the under-represented or missing classes,
an auxiliary source domain is required.
In Fig. \ref{fig:ihd2_target_cm}, we compare the confusion matrices of pseudo-labels 
from $4$ different models.
Significant improvements of pseudo-label accuracy can be observed 
after introducing auxiliary source domain \ihdone coupled with the proposed DA-AIM.
\input{figures/sup-figtex/FIG_conf_mat_ihd2_target}

%
\section{Qualitative Results}
Fig. \ref{fig:qualitatives-ihd2} and \ref{fig:qualitatives-ihd1} present 
qualitative DA action detection results of our proposed model trained on UDA protocols 
\avakins+\ihdone $\to$ \ihdtwo and \avakins+\ihdtwo $\to$ \ihdone respectively.
Sample detection results are shown on the respective target domain's validation frames.
Note that, the proposed DA-AIM model
can successfully detect action classes such as ``touch'', ``throw'', ``take-a-photo'',  ``stand'' and ``walk''.



    
%

%% file: figures/sup-figtex/FIG_conf_mat_avakin2ihd2_3.tex
\begin{figure}[t]
\centering
\includegraphics[width=0.6\linewidth]{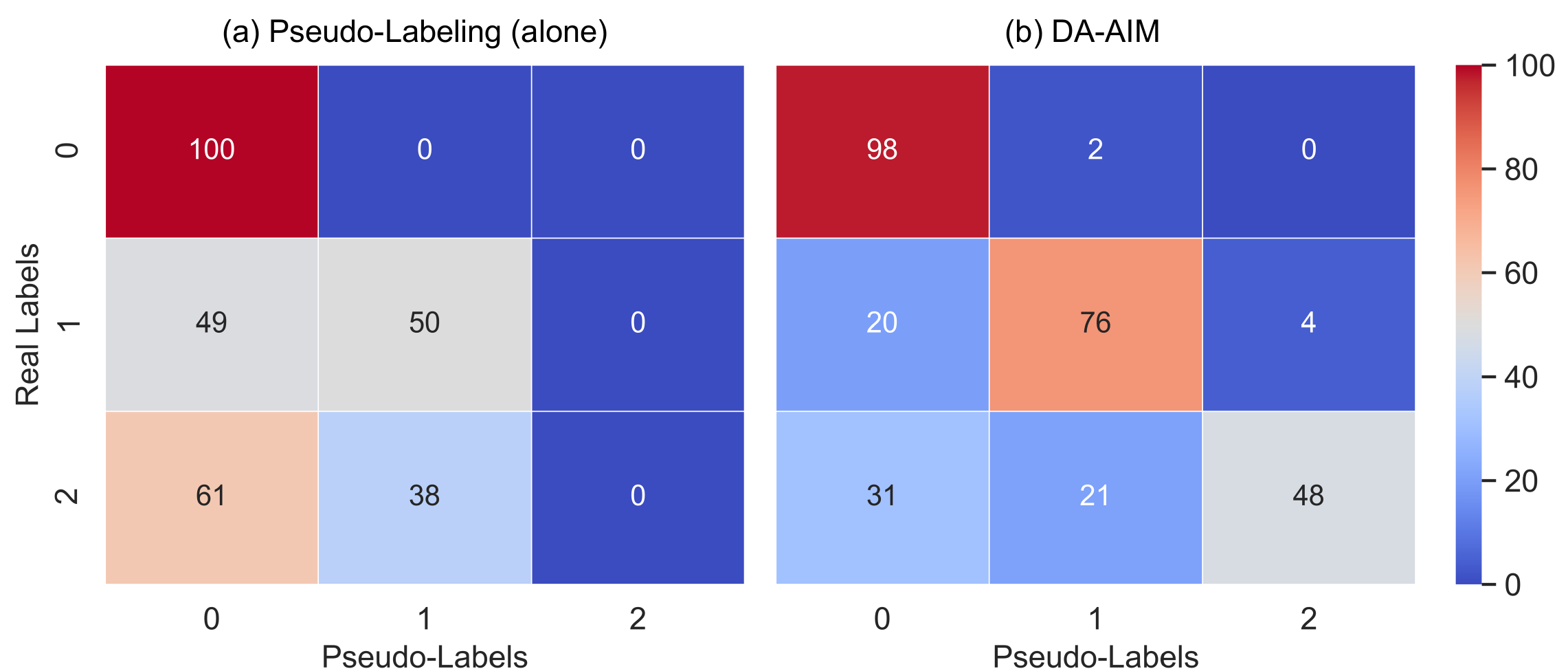}

\caption{\small{Comparison of confusion matrices (real-labels vs. pseudo-labels) of two different UDA approaches.
These two models are trained on \avakin$\to$\ihdtwo.
The first model (a) is trained using only pseudo-labels without the proposed adaptation approach.
The second model (b) is trained following the proposed DA-AIM approach.
Note, our proposed DA-AIM helps improving the quality of the pseudo labels.
}
}
\label{fig:avakin2ihd2_cm}
\end{figure}

%% file: figures/sup-figtex/FIG_conf_mat_ihd2_target.tex
\begin{figure*}[t]
\centering
\includegraphics[width=0.7\linewidth]{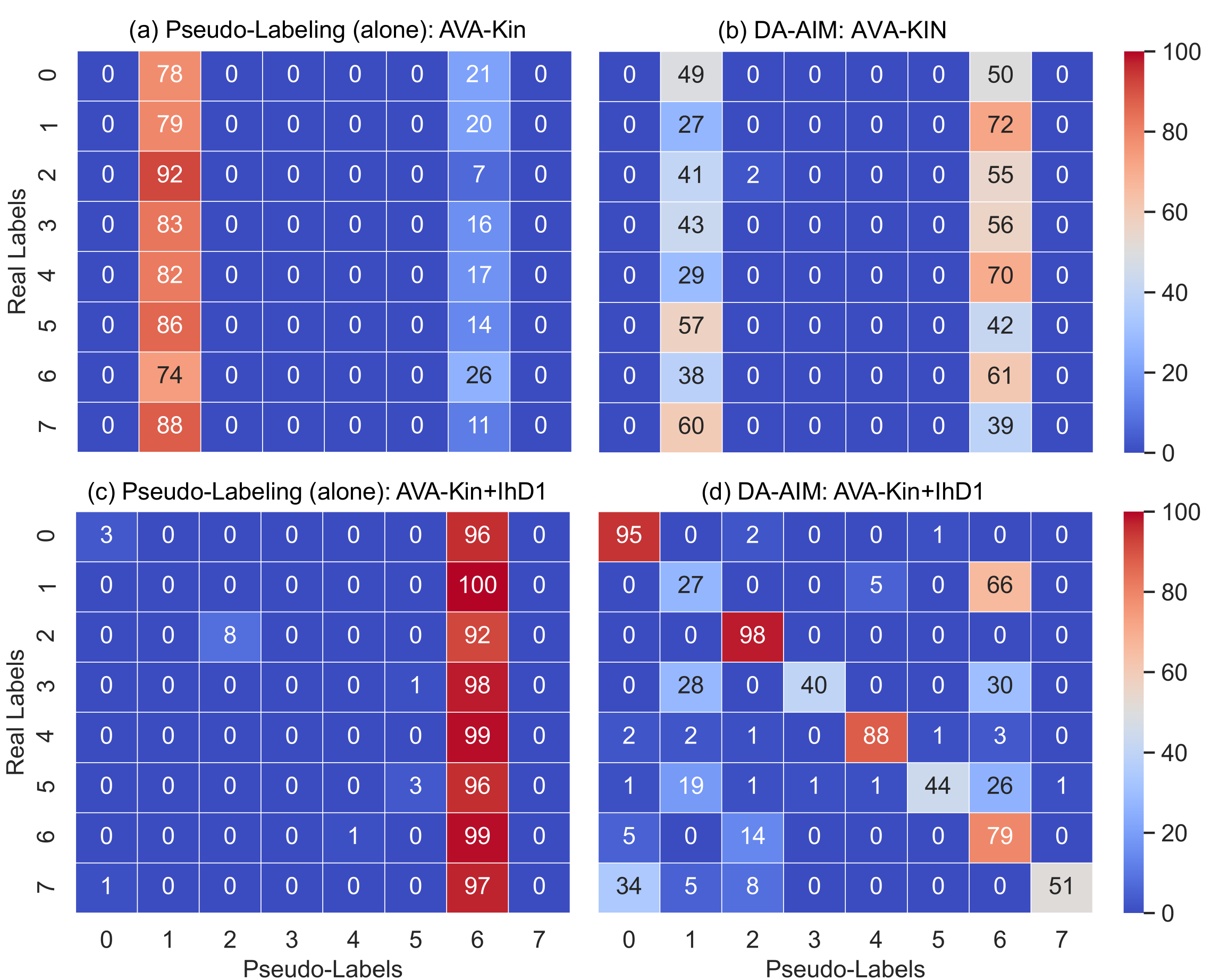}

\caption{\small{
Comparison of confusion matrices (real-labels vs. pseudo-labels) of $4$ different models trained on $8$ action classes.
Models are trained following either \avakin $\to$ \ihdtwo (a,b); 
or \avakin$+$\ihdone $\to$ \ihdtwo (c,d).
Note that best quality pseudo-labels are achieved when we perform 
auxiliary source domain based adaptation.  
}.
}
\label{fig:ihd2_target_cm}
\end{figure*}